\documentclass[lettersize,journal]{IEEEtran}
\usepackage{amsmath,amsfonts}
\usepackage{algorithmic}
\usepackage{algorithm}
\usepackage{array}
\usepackage[caption=false,font=normalsize,labelfont=sf,textfont=sf]{subfig}
\usepackage[colorlinks=true, linkcolor=blue, citecolor=blue]{hyperref}
\usepackage{textcomp}
\usepackage{stfloats}
\usepackage{url}
\usepackage{verbatim}
\usepackage{graphicx}
\usepackage{booktabs}
\usepackage{xcolor}
\usepackage{multirow}
\usepackage{pifont}
\usepackage{makecell}
\makeatletter

\newcommand{\Rmnum}[1]{\expandafter\@slowromancap\romannumeral #1@}
\makeatother
\usepackage{colortbl}

\begin{document}

\title{ARFC-WAHNet: Adaptive Receptive Field Convolution and Wavelet-Attentive Hierarchical Network for Infrared Small Target Detection}


\author{Xingye Cui, Junhai Luo,~\IEEEmembership{Member,~IEEE,} Jiakun Deng, Kexuan Li, Xiangyu Qiu and Zhenming Peng,~\IEEEmembership{Member,~IEEE,}
\thanks{Manuscript received XXXXX 00, 0000; revised XXXXX 00, 0000; accepted XXXXX 00, 0000.This work was supported by Natural Science Foundation of Sichuan Province of China (Grant No.2023NSFSC0508, 2025ZNSFSC0522) and partially supported by National Natural Science Foundation of China (Grant No.61571096). (Corresponding author: Junhai Luo). }
\thanks{The authors are with the School of Information and Communication Engineering Chengdu 611731, China, also with the Laboratory of Imaging Detection and Intelligent Perception, University of Electronic Science and Technology of China, Chengdu 610054, China. (e-mail: cxy011211@163.com; junhai\_luo@uestc.edu.cn; dengjiakun@std.uestc.edu.cn; kexuanli@std.uestc.edu.cn; 202421011422@std.uestc.edu.cn; zmpeng@uestc.edu.cn).}}

\maketitle
\begin{abstract}
Infrared small target detection (ISTD) is critical in both civilian and military applications. However, the limited texture and structural information in infrared images makes accurate detection particularly challenging. Although recent deep learning-based methods have improved performance, their use of conventional convolution kernels limits adaptability to complex scenes and diverse targets. Moreover, pooling operations often cause feature loss and insufficient exploitation of image information. To address these issues, we propose an adaptive receptive field convolution and wavelet-attentive hierarchical network for infrared small target detection (ARFC-WAHNet). This network incorporates a multi-receptive field feature interaction convolution (MRFFIConv) module to adaptively extract discriminative features by integrating multiple convolutional branches with a gated unit. A wavelet frequency enhancement downsampling (WFED) module leverages Haar wavelet transform and frequency-domain reconstruction to enhance target features and suppress background noise. Additionally, we introduce a high-low feature fusion (HLFF) module for integrating low-level details with high-level semantics, and a global median enhancement attention (GMEA) module to improve feature diversity and expressiveness via global attention. Experiments on public datasets SIRST, NUDT-SIRST, and IRSTD-1k demonstrate that ARFC-WAHNet outperforms recent state-of-the-art methods in both detection accuracy and robustness, particularly under complex backgrounds. The code is available at https://github.com/Leaf2001/ARFC-WAHNet.
\end{abstract}


\section{Introduction}
\IEEEPARstart{I}{nfrared} small target detection (ISTD) is crucial in both civilian and military. Unlike visible light small target detection, ISTD exhibits strong robustness under all-weather interference, making it widely applicable in areas such as maritime rescue, precision guidance, and fire alarm systems \cite{r1}, \cite{r2}. However, detecting infrared small targets remains challenging due to their inherent properties: 1) \(Small\): targets often occupy less than 0.15\% of image pixels and lack clear shape or texture cues; 2) \(Dim\): low contrast and signal-to-noise ratio (SNR) make them easily confused with background clutter \cite{r3}; 3) \(Size\) \(varying\): target size changes with distance and imaging conditions. These factors make ISTD a persistent and complex research challenge.

In recent decades, ISTD has undergone significant evolution, primarily categorized into model-driven and data-driven methods. Traditional model-driven methods- based on background consistency \cite{r4}, \cite{r5}, sparse representation \cite{r6}, \cite{r7}, and human visual system (HVS) assumptions \cite{r8}, \cite{r9}, \cite{r10}, perform well in simple scenes but suffer from limited generalization due to parameter sensitivity and dependence on prior knowledge. Furthermore, these methods exhibit three critical limitations: 1) poor robustness in cluttered scenes, 2) lack of real-time capability, and 3) high computational cost, hindering practical deployment. To address these constraints, recent research has shifted toward data-driven methods \cite{r11}, \cite{r12}, \cite{r13}. Convolutional neural networks (CNNs) effectively capture deep semantic features, while generative adversarial networks (GANs) \cite{r14} have been introduced to balance detection accuracy and false alarm rates.  These methods leverage autonomous feature learning, offering enhanced robustness for ISTD in complex scenes.

Despite the success of data-driven methods, two key challenges remain in ISTD: 1) Limited expressiveness of fixed convolutions. Most existing methods rely on static convolutional kernels that struggle to adapt to target variations under diverse imaging conditions \cite{r15}, \cite{r16}, \cite{r17}. Although some approaches introduce handcrafted priors \cite{r18}, \cite{r19}, \cite{r20}, they are often limited to simple concatenations and fail to leverage the full potential of dynamic and self-adaptive convolutions. Consequently, developing dynamic convolution selection could empower networks to more effectively handle heterogeneous infrared scenarios. 2) Inadequate feature preservation in hierarchical networks. The inherent deficiency of distinct texture and shape-related fine-grained features in infrared small targets exacerbates information loss in existing deep architectures. Conventional networks capture deep semantic information through progressive downsampling of feature maps \cite{r18}, which may cause target submergence and excessive detail loss. Therefore, improved spatial precision and feature retention mechanisms are essential for preserving critical target signatures.

To address the above challenges, we propose ARFC-WAHNet: an adaptive receptive field convolution and wavelet-attentive hierarchical network for infrared small target detection. Specifically, the multi-receptive field feature interaction convolution (MRFFIConv) module replaces standard 3×3 convolutions in the encoder-decoder framework to enhance multi-scale feature extraction, morphological adaptability, and directional edge perception. It integrates multi-scale dilated convolution (MSDC) \cite{r21}, deformable convolution (DCN) \cite{r22}, and multi-directional difference convolution (MDDC), coupled with a gated unit that dynamically selects processing paths based on target–background variations. To reduce target loss during downsampling, we introduce the wavelet frequency enhancement downsampling (WFED) module in the encoder. By combining Haar wavelet transform (HWT) with frequency-domain enhancement, WFED significantly improves small target feature representation.

\begin{figure}[t]
	\centering 
	\includegraphics[width=0.5\textwidth]{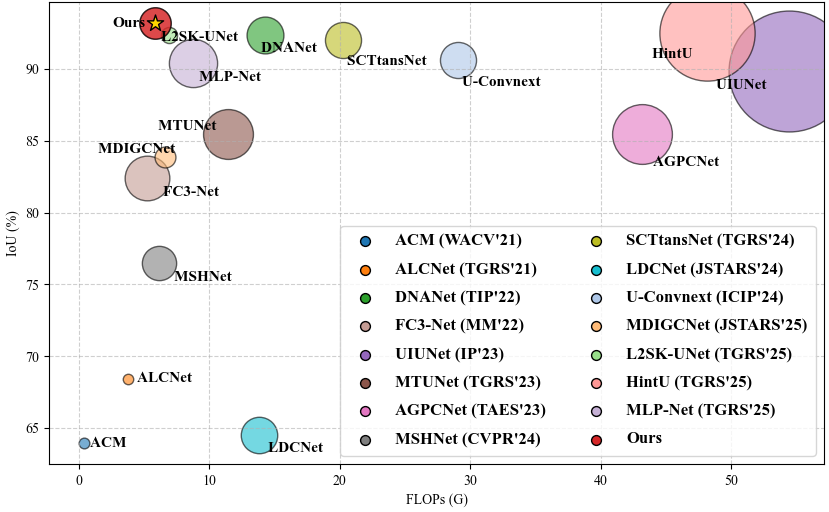}
	\caption{Comparison of IoU, parameter count (\#Params.), and FLOPs of mainstream ISTD deep learning methods on the NUDT-SIRST dataset \cite{r35}.} 
 \label{Fig.1}
\end{figure}

Furthermore, the high-low feature fusion (HLFF) module fuses low-level detail with high-level semantics, enhancing resolution and localization. The global median enhancement attention (GMEA) module leverages global statistics and multi-scale spatial attention to improve feature representation. As shown in Fig. \ref{Fig.1}, ARFC-WAHNet achieves superior detection performance with fewer FLOPs and parameters than mainstream state-of-the-art (SOTA) backbones. The main
contributions of this article are as follows.

1) Inspired by dynamic convolution, we propose MRFFIConv by integrating parallel expert convolution branches with a gating unit to address the diverse environmental complexity and small target characteristics across different scenarios.

2) We propose WFED for downsampling, which enhances edges and details while suppressing noise and background information, effectively mitigating feature loss in hierarchical networks.

3) To further enhance information utilization, we design HLFF to fuse low-level details with high-level semantics and introduce GMEA to enable inter-channel interaction, strengthening multi-scale feature integration.

4) The proposed network achieves SOTA performance across multiple datasets. Visualization results further demonstrate its superior detection accuracy and robustness in various challenging scenarios.

The remainder of this article is organized as follows: Section II reviews related work in recent years. Section III details the ARFC-WAHNet architecture and its key modules. Section IV presents ablation and comparative experiments to validate the proposed method. Section V concludes the paper.

\begin{figure*}[t]
	\centering 
	\includegraphics[width=\textwidth]{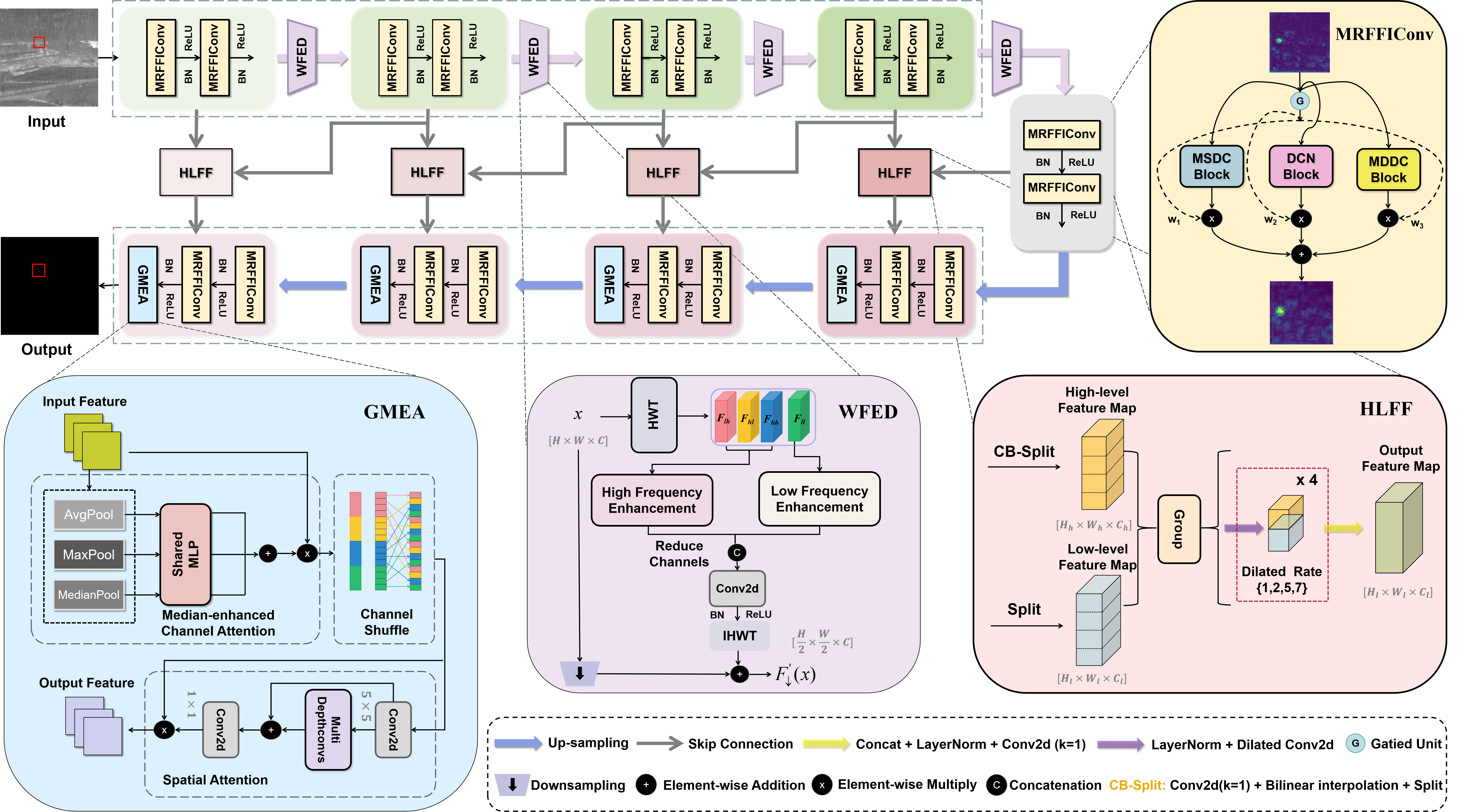}
	\caption{The overall architecture of the proposed ARFC-WAHNet, which has a U-Net framework with MRFFConv, WFED, HLFF, and GMEA modules.} 
 \label{Fig.2}
\end{figure*}

\section{RELATED WORK}
\subsection{Infrared Small Target Detection}
In general, ISTD algorithms can be divided into model-driven and data-driven methods.
\subsubsection{Model-Driven Methods}Model-driven methods for ISTD mainly include those based on background consistency assumptions, sparse representation, and HVS mechanisms. Background consistency-based methods assume strong correlation in the background regions of infrared images. For instance, maximum mean and maximum median filters \cite{r4} preserve target edges while suppressing clutter, and the Tophat algorithm \cite{r5} enhances target peaks while attenuating background interference. However, in complex backgrounds, these methods are prone to noise and clutter, resulting in high false alarm rates. Sparse representation-based methods exploit the non-local self-similarity of the background and the sparsity of targets by formulating detection as a low-rank and sparse matrix recovery problem. IPI \cite{r6} pioneered this approach by modeling background patches as low-rank components and targets as sparse outliers. Several enhanced variants followed \cite{r23}, \cite{r24}, \cite{r25}. To improve efficiency, Dai et al. \cite{r26} proposed RIPT, which integrates local and non-local priors, while PSTNN \cite{r7} reduces computational complexity via reweighted sparsity and tensor decomposition. Nevertheless, these methods struggle with dim, contourless targets and often lack real-time performance. HVS-based methods rely on local saliency to distinguish targets from clutter. Chen et al. introduced the LCM \cite{r27} to capture contrast between a pixel and its neighborhood. Subsequent works refined contrast computation to improve robustness. For example, Wei et al. \cite{r8} proposed MPCM using multi-scale windows, and Han et al. \cite{r9} extended this to a multi-layer structure to adapt to varying target sizes. Further developments \cite{r29}, \cite{r30}, \cite{r31} improved computational efficiency. However, due to reliance on hand-crafted shallow features, these methods generally exhibit limited generalization across diverse scenarios.

\subsubsection{Data-Driven Methods}With the rapid development of deep learning, the limitations of traditional ISTD methods have been gradually alleviated, and data-driven approaches have opened new avenues for infrared small target detection. These methods typically generate saliency maps by designing specialized network architectures that extract discriminative features from infrared imagery and perform target segmentation for accurate detection. Dai et al. proposed the ACM module \cite{r18}, which can be embedded into existing frameworks to replace traditional feature fusion schemes. They also introduced ALCNet \cite{r32}, a deep, parameter-free module reformulated from local contrast-based methods to enhance small target representation. Attention mechanisms have also been widely integrated into ISTD networks. For example, LDCNet \cite{r43} employs differential attention in a cascade structure to enhance target features, while MDIGCNet \cite{r45} introduces a multi-directional context-aware module to improve focus on small targets. To address scale variation challenges, various feature fusion strategies have been proposed \cite{r16}, \cite{r33}, \cite{r41} and network architectures have been further optimized \cite{r37}, \cite{r38}, \cite{r39}, \cite{r40}, \cite{r44} to capture multi-scale, global, and positional information more effectively. Additionally, to compensate for the low information content in infrared imagery, some works utilize densely nested attention networks \cite{r35} or cross-layer correlation analysis \cite{r36}. Learnable kernel functions \cite{r42} have also been explored to guide saliency detection and enhance small target prominence and fusion. Overall, data-driven methods have demonstrated significant advantages in ISTD across various complex scenes.

\subsection{Dynamic Convolution}
Dynamic convolution allows adaptive adjustment of model parameters based on input data and has been widely adopted in computer vision tasks. Inspired by the mixture of experts framework \cite{r46}, Yang et al. introduced CondConv \cite{r47}, which models convolution as a multi-branch operation, with each branch corresponding to an individual kernel. Building on this, DY-CNN \cite{r48} employs an attention mechanism to weight multiple kernels, enabling dynamic feature extraction. Brabandere et al. \cite{r49} proposed input-dependent filters for tasks like next-frame and stereo prediction, while Dai et al. \cite{r50} learned sample-specific convolutional offsets. In graph domains, ECC \cite{r51} generates filter weights based on edge attributes, capturing both local and global patterns efficiently.

Recognizing the potential of dynamic convolution in ISTD, Nian \cite{r52} proposed ParC-DPConv, which predicts kernel parameters conditioned on input features to enhance generalization. DRPN \cite{r53}, introduced by Peng et al., incorporates multi-branch convolutions of varying sizes with a dynamic re-parameterization strategy to address scale variation in multi-frame infrared targets. However, these methods still struggle to adapt to the diverse characteristics of small targets and the complexity of environments across different scenes. Therefore, further advancements in dynamic convolution are needed to improve model adaptability in ISTD tasks.

\section{METHODOLOGY}
\subsection{Overall Architecture}
The proposed ARFC-WAHNet consists of five core components: a backbone network, the MRFFIConv module, the WFED module, the HLFF module, and the GMEA module, as illustrated in Fig. \ref{Fig.2}. Built upon a symmetric encoder-decoder architecture \cite{r28}, the backbone gradually extracts multi-level features via the encoder and restores spatial information through the decoder. Skip connections are incorporated to fuse deep semantic and shallow spatial information, facilitating the detection of weak and small targets.

The encoder-decoder consists of five stages, with MRFFIConv replacing standard convolutions to enhance feature extraction for ISTD. It uses parallel convolutional branches with varying receptive fields and a dynamic gated unit, improving multi-scale feature extraction, morphological adaptability, and edge perception. The gated fusion of branches balances performance gains with computational efficiency.

The WFED module enhances small target features by decomposing input features via HWT into low-frequency and high-frequency subbands. It emphasizes high-frequency details and suppresses background noise using adaptive filtering. Enhanced features are fused and reconstructed through inverse wavelet transform, with residual connections preserving context and reducing information loss.

To enhance cross-level feature fusion, HLFF improves skip connections by merging low- and high-level features. It splits features into four channel groups, applies varying dilation rates, and fuses them via a 1×1 convolution. This boosts the network’s ability to localize and identify small targets.

Finally, the GMEA module integrates spatial and channel attention by combining channel shuffling, median pooling for global statistics, and multi-scale convolutions, thereby enriching feature representation and boosting detection performance.

\begin{figure}[t]
	\centering 
	\includegraphics[width=0.49\textwidth]{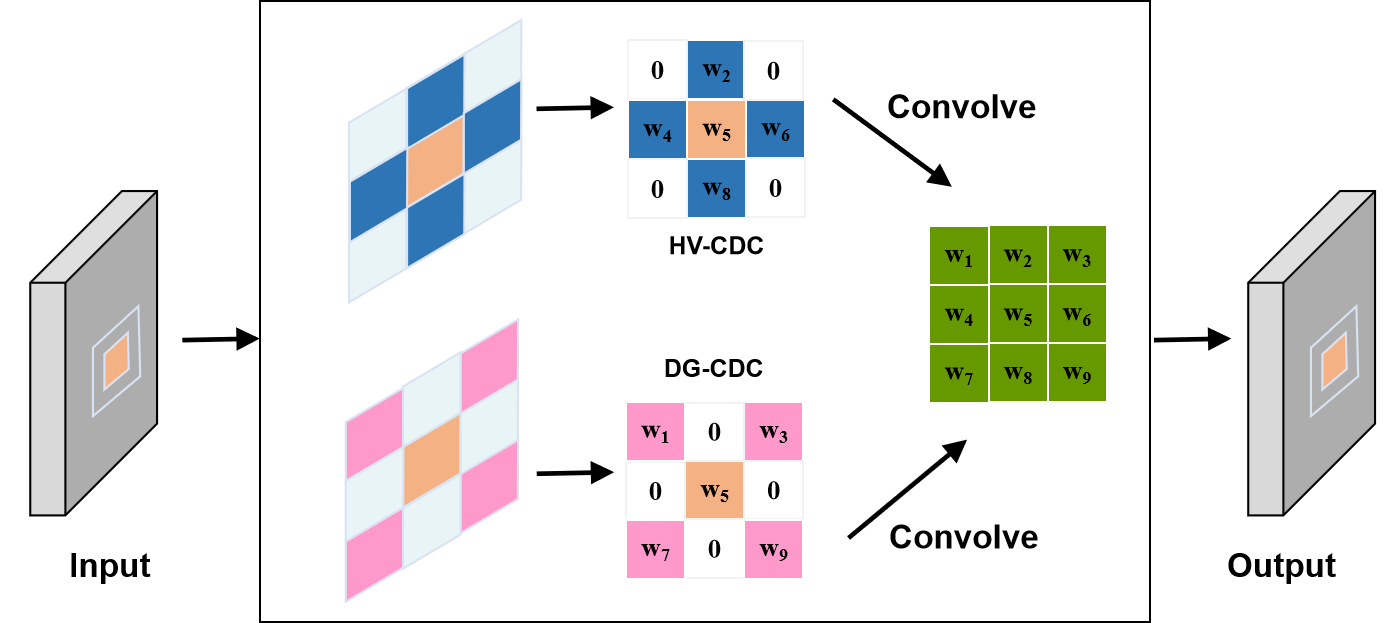}
	\caption{The structure of horizontal and vertical central difference convolution (HV-CDC) and diagonal central difference convolution (DG-CDC).} 
 \label{Fig.3}
\end{figure}\textbf{}

\begin{figure}[t]
	\centering 
	\includegraphics[width=0.49\textwidth]{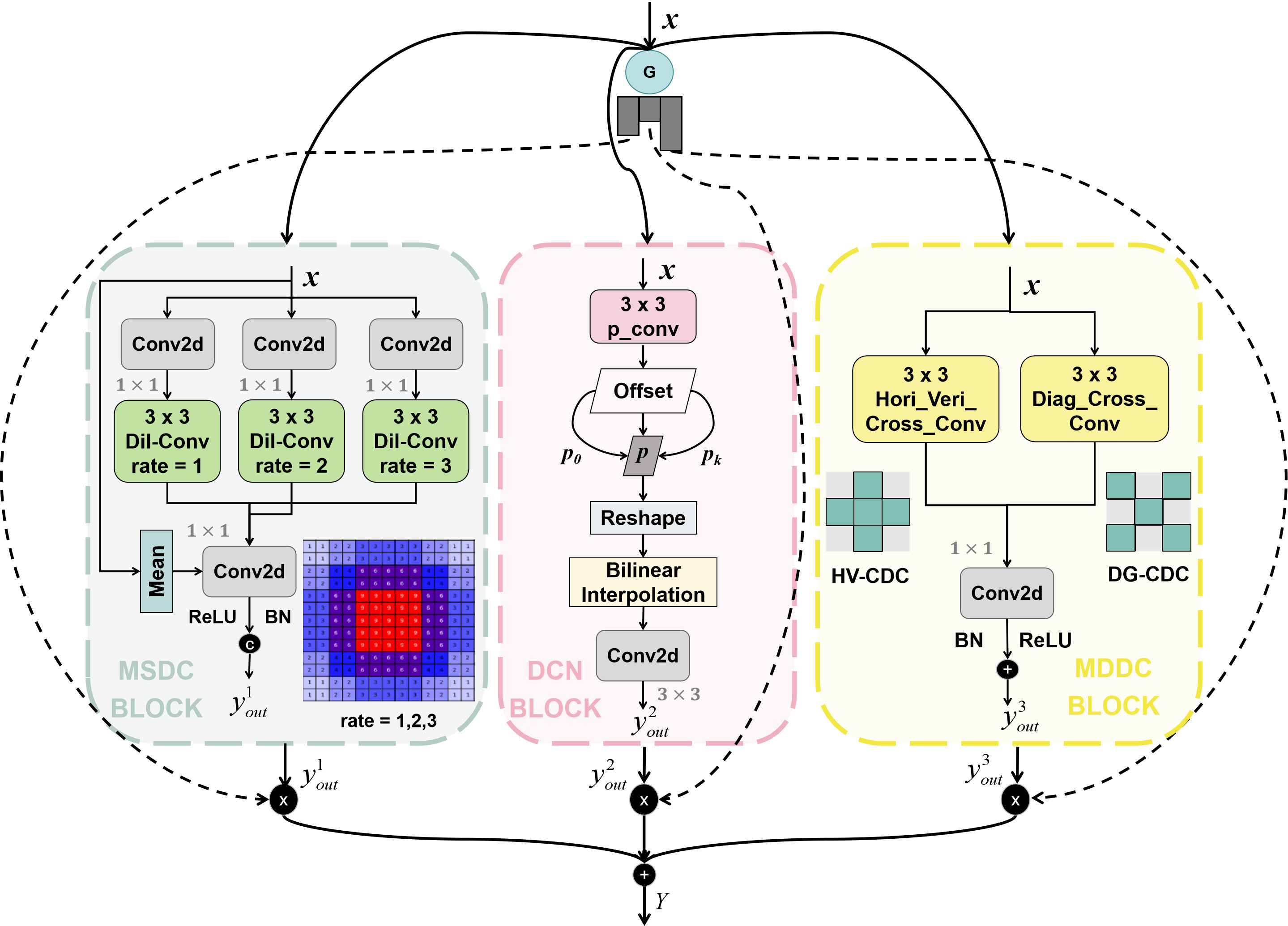}
	\caption{Illustration of the MRFFIConv module.} 
 \label{Fig.4}
\end{figure}\textbf{}

\subsection{MRFFIConv Module}
ISTD remains challenging due to the intricate interplay between weak targets and complex background clutter. Conventional convolutional layers apply identical kernels across all inputs, limiting their adaptability and often resulting in suboptimal performance in heterogeneous scenarios.

Inspired by existing works \cite{r47}, \cite{r53}, we introduce the MRFFIConv module, which integrates a gated unit and three convolutional branches: MSDC, DCN, and MDDC. Each branch is tailored for distinct subtasks-multi-scale feature extraction, shape adaptability, and edge-detail enhancement, respectively.

\subsubsection{MSDC}Compared to standard convolution, the MSDC branch employs dilated convolutions with rates \(\left\lbrace1,2,3\right\rbrace\) to expand the receptive field of each pixel, enabling more effective integration of local details and global context for multi-scale feature extraction. Global average pooling is also incorporated to capture holistic contextual information. The outputs are concatenated to form a multi-scale representation. To enhance training stability and efficiency, batch normalization (BN) and rectified linear unit (ReLU) are applied after each convolution. The convolution operation in each branch can be expressed as
\begin{equation}
\label{deqn_ex1}
{y}_{k}=\mathrm{ReLU}(\mathrm{BN}(\mathrm{Conv}^{d_k}_\mathrm{3x3}(x,w_{k},d_{k})))
\end{equation}
where \(x\) is the input feature map, \(y_k\) indicates the output feature map of the \(k\)-th branch, \(d_k\) represents the dilation rate of the \(k\)-th branch (\(d_1=1\), \(d_2=2\), \(d_3=3\)), \(\mathrm{Conv}^{d_k}_\mathrm{3x3}\) is a 3×3 convolution operation with a dilation rate of \(d_k\), and \(w_k\) stands for the convolution kernel weight of the \(k\)-th branch.

We express the formula for the entire output \(y^1_{out}\) as
\begin{equation}
\label{deqn_ex2}
y^1_{out} =\mathrm{ReLU}(\mathrm{BN}(w_{\mathrm{Cat}}\cdot{\mathrm{Cat}}({y}_{k},G_{\mathrm{out}})))
\end{equation}
where \(w_{\mathrm{Cat}}\) is the weight of the 1x1 convolution, and \(G_{\mathrm{out}}\) represents the output of the global feature extraction branch. \(\mathrm{Cat}\) is the concatenation operation.

\subsubsection{DCN}By introducing long-range dependencies and adaptive spatial aggregation in convolution, DCN can further enhance the geometric transformation modeling capability of the entire network. DCN can be formulated as
\begin{equation}
\label{deqn_ex3}
y^2_{out}=\sum_{k=1}^Kw_km_kx(p_0+p_k+{\Delta}p_k)
\end{equation}
where \(x\) is the input feature map with a shape of \((H,W,C)\), \(K\) represents the total number of sampling points, and k represents each sampling point. \(w_k\in {R}^{C \times C}\) denotes the projection weight of the \(k\)-th sampling point, while \(\quad m_k \in R\) represents the modulation scalar of the \(k\)-th sampling point, which is normalized using the sigmoid function. \(p_0\) is the current pixel, \(p_k\) indicates the \(k\)-th position in the predefined grid sampling, and \(\Delta p_k\) represents the offset of the \(k\)-th grid sampling position.

Due to the input-dependent sampling offset \(\Delta p_k\) and modulation scalar \(m_k\), this adaptability allows the DCN to better address the challenges of processing complex scenes in infrared images. For example, the model adaptively refines sampling positions and weights to better capture target features in the presence of strong noise.

\subsubsection{MDDC}Apart from the common central difference convolution (CDC), we tend to sample a sparser local area, fully utilizing the local features and interactions between vertical-horizontal directions and diagonal views, and propose the MDDC. It can be represented as
\begin{equation}
\label{deqn_ex4}
y(p_0)=\sum_{p_n\in{\mathcal{S}}}w(p_n)\cdot(x(p_0+p_n)-x(p_0))
\end{equation}
where \(p_n\) enumerates locations of the local critical region \(\mathcal{S}\).

Specifically, we decouple \(\mathcal{S}\) into two intersecting adjacent regions to explicitly learn structural differences in the feature maps along different orientations, including 1) horizontal and vertical (HV) intersecting adjacent regions \(\mathcal{S}_{HV}= \{(-1,0), (0,-1), (0,0), (0,1), (1,0)\}\); and 2) diagonal (DG) intersecting neighbor regions \(\mathcal{S}_{DG} = \{(-1,-1), (-1,1), (0,0), (1,-1), (1,1)\}\). Thus, when \(\mathcal{S} = \mathcal{S}_{HV}\) and \(\mathcal{S} = \mathcal{S}_{DG}\), they can represent horizontal and vertical central difference convolution (HV-CDC) and diagonal central difference convolution (DG-CDC), respectively. Fig.\ref{Fig.3} illustrates the workflow of HV-CDC and DG-CDC. The final output \( y^3_{out} \) of the MDDC Block is obtained by element-wise summing the feature maps from both convolutions. Parallel deployment of these convolutions boosts the model’s ability to distinguish infrared small targets from clutter.

\subsubsection{Gated Unit}In the MRFFIConv module, the convolutional kernel is computed as a function of the input example. Specifically, we parameterize these convolutions by
\begin{equation}
\label{deqn_ex5}
Y(x;\theta,\{W_{i}\}_{i=1}^{3})=\sum_{i=1}^{3}G(x,\theta)_{i}y^i_{out}(x;W_{i})
\end{equation}
where \(Y\) denotes the overall model output, \(x\) represents the input, \(\theta\) corresponds to global parameters, \(\{W_{i}\}_{i=1}^{3}\) indicates the parameters of the \(i\)-th convolutional branch, and \(y^i_{out}\) is the output of each convolutional block.


We aim to design a computationally efficient discriminative function that ensures effective separability between heterogeneous inputs. Features are extracted via the activation function, scored by a fully connected layer, and normalized using softmax to produce a probability distribution, ensuring branch weights sum to one. This process is formulated as
\begin{equation}
\label{deqn_ex6}
G(x,\theta)_{i}=\mathrm{Softmax}{(g(x;\theta))}_{i}=\frac{\mathrm{exp}((g(x;\theta))_{i})}{\sum_{j=1}^{3}{(g(x;\theta))_{i}}}
\end{equation}
where \(G(x,\theta)_{i}\) is the normalized weight and \((g(x;\theta))_{i}\) the raw gating score of the \(i\)-th convolutional branch. Fig. \ref{Fig.4} reflects the design of MRFFIConv and its connection to conditional computation and mixture-of-experts models.

\begin{figure}[t]
	\centering 
	\includegraphics[width=0.49\textwidth]{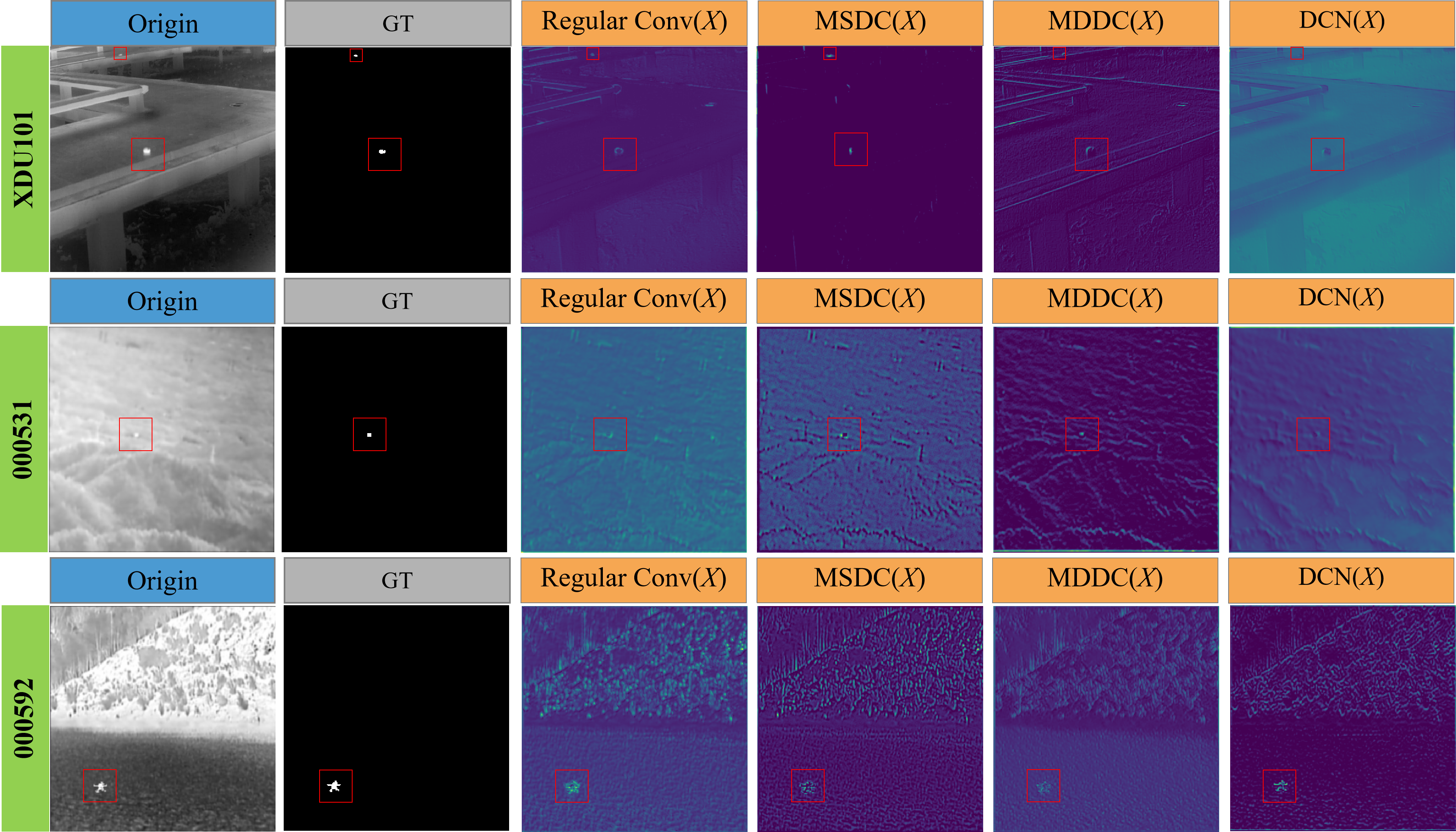}
	\caption{Illustration of a heatmap. The columns from left to right represent the original image, ground truth, heatmap output from Regular Conv, MSDC, MDDC, and DCN, respectively.} 
 \label{Fig.5}
\end{figure}

To evaluate expert capabilities, we test across three representative ISTD scenarios \cite{r4}, \cite{r55}: 1) urban scenes with varying target sizes, 2) small targets in dense clouds, and 3) drones over the sea. As shown in Fig. \ref{Fig.5}, MSDC captures multi-scale targets, DCN adapts to complex shapes, and MDDC preserves fine details. The gated unit dynamically selects expert paths based on input, enabling adaptive computation without added inference cost, offering an efficient solution for diverse ISTD challenges.

\textbf{}

\subsection{WFED Module}
\begin{figure}[t]
	\centering 
	\includegraphics[width=0.49\textwidth]{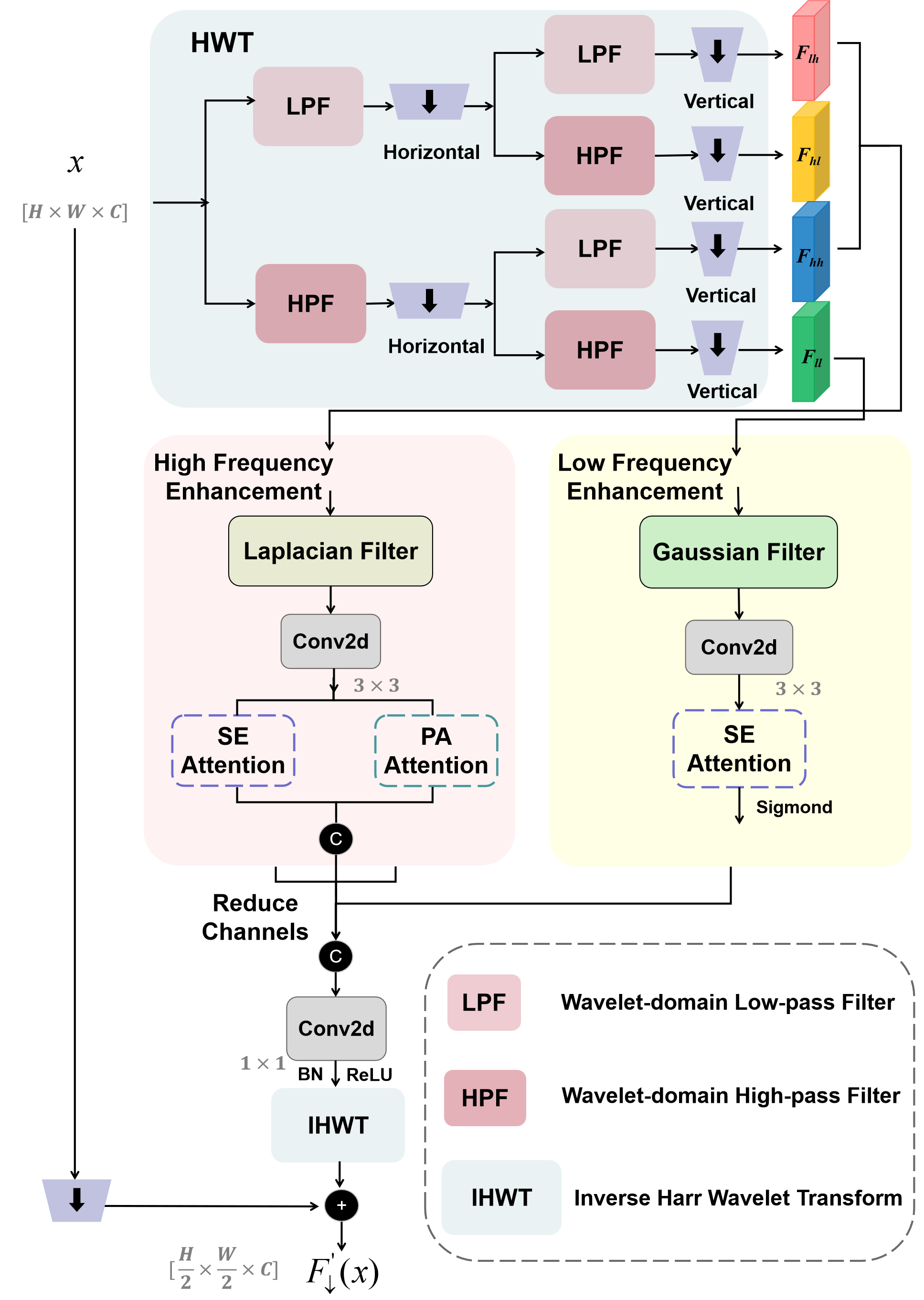}
	\caption{Illustration of the WFED module.} 
 \label{Fig.6}
\end{figure}

Conventional downsampling methods often result in significant information loss, which is especially detrimental in tasks like ISTD \cite{r8}, \cite{r19}. We propose an innovative method WFED module, as shown in Fig. \ref{Fig.6}. This module utilizes 2-D Haar wavelet transform (2D-Haar-WT) and frequency enhancement downsampling to replace the traditional max pooling operation, thereby relieveing information loss in ISTD.

The original feature map \(F\) first undergoes a convolution operation to extract preliminary features. It is then processed by the 2D-Haar-WT, which sequentially applies wavelet-domain low-pass filter (LPF) and high-pass filter (HPF) along the horizontal and vertical directions, followed by downsampling by a factor of 2 in each dimension. This operation decomposes the feature map into four sub-band components: \({F_{ll},F_{lh},F_{hl},F_{hh}}\), where each has one-fourth the original number of channels and half the spatial resolution. Specifically, \(F_{ll}\) is the low-low (approximation) coefficient, \({F_{lh}}\) is the low-high (horizontal detail) coefficient, \(F_{hl}\) is the high-low (vertical detail) coefficient, and \(F_{hh}\) is the high-high (diagonal detail) coefficient. The mathematical representation of this transformation is

\begin{equation}
\label{deqn_ex7}
\{F_{ll},\,F_{hl},\,F_{lh},\,F_{hh}\} = \mathrm{HWT}(f(F))
\end{equation}
where \(f(\cdot)\) represents the convolutional operation and \(\mathrm{HWT}(\cdot)\) represents the 2D-Haar-WT.
Specifically, after then, the value of \(F_{ll}\) at position \((i,j)\) is calculated as follows
\label{deqn_ex8}
\begin{align}
\begin{split}
F_{ll}(i,j) &= F(2i-1,2j-1) + F(2i-1,2j) \\
&\quad + F(2i,2j-1) + F(2i,2j)
\end{split}
\end{align}

\(F_{ll}\) captures the low-frequency components, effectively retaining the primary structural features of the image, while \({F_{lh},F_{hl},F_{hh}}\) represent high-frequency components containing texture and edge details. These sub-bands are utilized in two forms: \(F^1\), which concatenates both low- and high-frequency components for further convolutional processing; and 
\(F^2\), which directly uses \(F_{ll}\) as input to the next network layer. This process can be formulated as follows
\begin{equation}
\label{deqn_ex9}
F^1 = \mathrm{Cat}(F_{ll},F_{lh},F_{hl},F_{hh})
\end{equation}
\begin{equation}
\label{deqn_ex10}
F^2 = F_{ll}
\end{equation}

To enhance local structural features, the three directional high-frequency sub-bands are further processed using a Laplacian filter. The frequency response of the employed frequency-domain high-pass filter is defined as
\begin{equation}
\label{deqn_ex11}
H(u,v)=\left\{
\begin{array}{l l}
	{{1-\exp\left(-\frac{D_{0}^{2}}{D^{2}(u,v)}\right)}} & {{\mathrm{if}\;D(u,v)\geq D_{0}},} \\
	{{0}}                                                & {{\mathrm{otherwise}}}
\end{array}\right.
\end{equation}
where \(D(u,v)=\sqrt{(u-u_0)^2+(v-v_0)^2}\), which denotes the radial distance from the spectral center in the frequency domain, and \(D_0 =max(H,W)//20\) represents the cutoff frequency that governs the attenuation threshold of the filter.

To augment the perception of high-frequency components, we incorporate the squeeze-and-excitation (SE) attention and pixel attention (PA). SE attention recalibrates global features to enhance salient target regions, while PA attention refines pixel-level responses, reinforcing the role of high-frequency features in ISTD. This process can be formulated as
\begin{equation}
\label{deqn_ex12}
F^E_h = \mathrm{Cat}(\mathrm{SE}(\mathcal{H}({F}_{lh/hl/hh})),\mathrm{PA}(\mathcal{H}({F}_{lh/hl/hh})))
\end{equation}
where \(F^E_h\) is the concatenated high-frequency feature map with channel dimensions reduced to three quarters of the original, \(\mathcal{H}(\cdot)\) is the frequency-
domain high-pass filtering operation, \(\mathrm{SE}(\cdot)\) and \(\mathrm{PA}(\cdot)\) denote the channel-wise and pixel-wise attention operations, respectively.

\textbf{}

For low-frequency processing, we apply a Gaussian filter to suppress high-frequency noise and enhance background smoothness. The frequency response of the applied filter is defined as
\begin{equation}
\label{deqn_ex13}
L(u,v)=\left\{
\begin{array}{l l}
	{{\exp\left(-\frac{D^{2}(u,v)}{D_{0}^{2}}\right)}} & {{\mathrm{if}\;D(u,v)\le D_{0}}}, \\
	{{1}}                                                & {{\mathrm{otherwise}}}
\end{array}\right.
\end{equation}

Furthermore, SE attention is combined with a sigmoid (\(\sigma\)) suppression module to adaptively reduce background redundancy, suppress interference, and enhance target contrast. This process can be formulated as
\begin{equation}
\label{deqn_ex14}
F^E_l = \mathrm{\sigma}(\mathrm{SE}(\mathcal{L}({F}_{ll}))
\end{equation}
where \(F^E_l\)  denotes the processed low-frequency components with channel dimensions reduced to a quarter of the original, and \(\mathcal{L}(\cdot)\) represents the frequency-domain low-pass filtering operation.

Subsequently, \(F^E_h\) and \(F^E_l\) are concatenated and fused through inverse Haar wavelet transform (IHWT) to reconstruct the enhanced features in the spatial domain. The reconstructed feature map can be formulated as
\begin{equation}
\label{deqn_ex15}
F^{E}=\mathrm{IHWT}(\mathrm{Cat}({F_{h}^{E}},{F_{l}^{E}}))
\end{equation}

Finally, we introduce a residual connection to obtain the fused features \(F^{'}_{\downarrow}(x)\) through
\begin{equation}
\label{deqn_ex16}
F^{'}_{\downarrow}(x)=F^{E}+\mathrm{Maxpool_{2x2}}(x)
\end{equation}
where the original input \(x\) undergoes 2×2 max pooling downsampling to match the spatial resolution of \(F^E\). This design preserves global contextual information while achieving frequency-domain enhancement, effectively alleviating information loss during downsampling.

\begin{figure}[t]
	\centering 
	\includegraphics[width=0.49\textwidth]{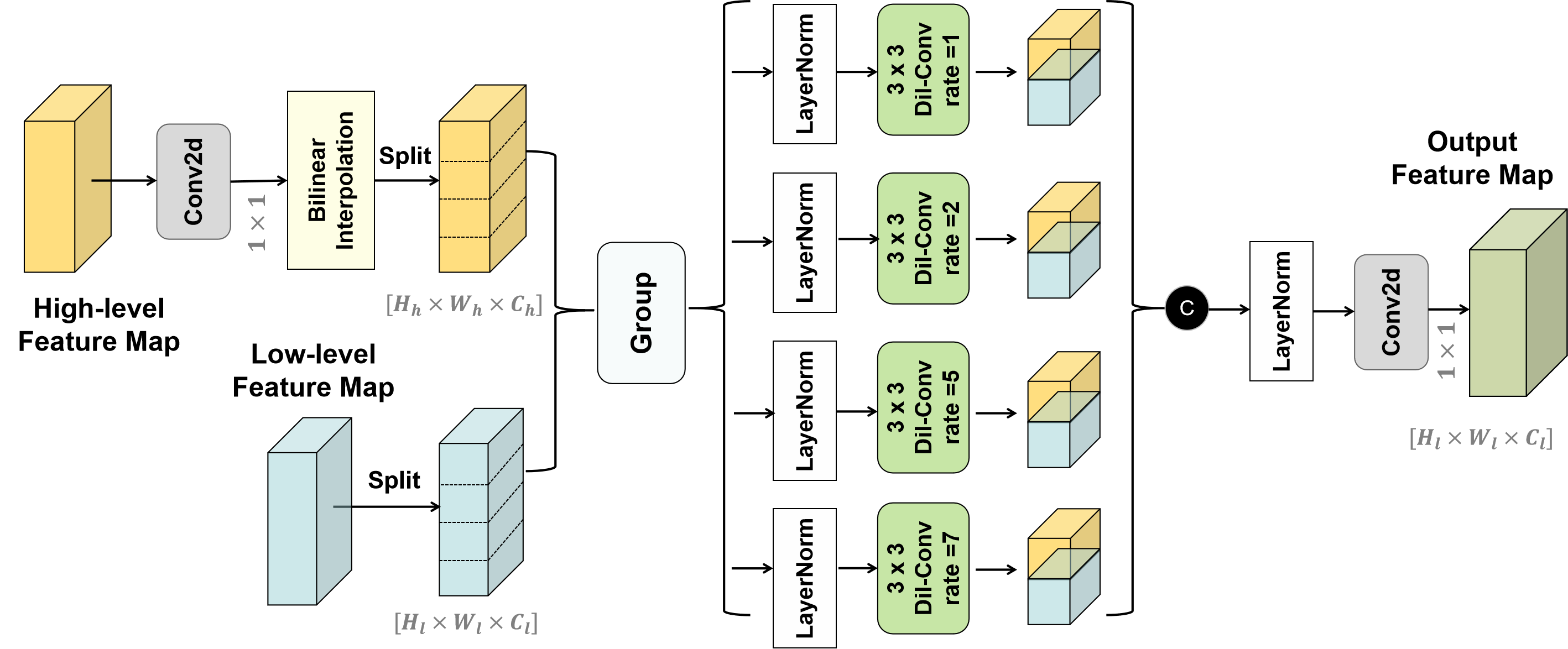}
	\caption{Illustration of the HLFF module.} 
 \label{Fig.7}
\end{figure}\textbf{}

\subsection{HLFF Module}
In CNNs, low-level features often contain rich detail information, while high-level features tend to contain more semantic information. In ISTD, small target features may be lost as the network deepens \cite{r33}, \cite{r35}. To mitigate this, we propose the HLFF module to fuse deep and shallow features, enhancing small target representation, as illustrated in Fig. \ref{Fig.7}.

First, we use depthwise separable convolution (DW) and bilinear interpolation (BI) to adjust the size of high-level features to match low-level features. Next, we divide the two feature maps along the channel dimension into four groups and connect one group of low-level features with one group of high-level features to obtain four sets of fused features. Dilated convolutions with rates \(\left\lbrace1,2,5,7\right\rbrace\) are applied to the fused groups to capture multi-scale information. Finally, all groups are concatenated along the channel dimension. This process is formulated as
\begin{equation}
\label{deqn_ex17}
x_{i}=\mathrm{Conv_{3\times3}}(\mathrm{LN}(\mathrm{Cat}(xh_i,xl_i)))
\end{equation}
\begin{equation}
\label{deqn_ex18}
x_{out}=\mathrm{Conv_{1\times1}}(\mathrm{LN}(\mathrm{Cat}(x_i)))
\end{equation}
where \(i\in\left\lbrace1,2,3,4\right\rbrace\), and \(xh_i\) is the high-level feature map obtained by adjusting the number of channels and spatial dimensions using DW and BI and grouping by channel dimension, and \(xl_i\) is the low-level feature map obtained by similarly grouping by channel dimension. This operation divides each input feature map into four parts. \(LN(\cdot)\) denotes layer normalization, and \(x_{out}\) is the final output.

Infrared images often exhibit significant target scale variation and complex backgrounds. Enhancing the model’s ability to localize and recognize small targets under such conditions improves detection accuracy and robustness.


\begin{figure*}[t]
	\centering 
	\includegraphics[width=\textwidth]{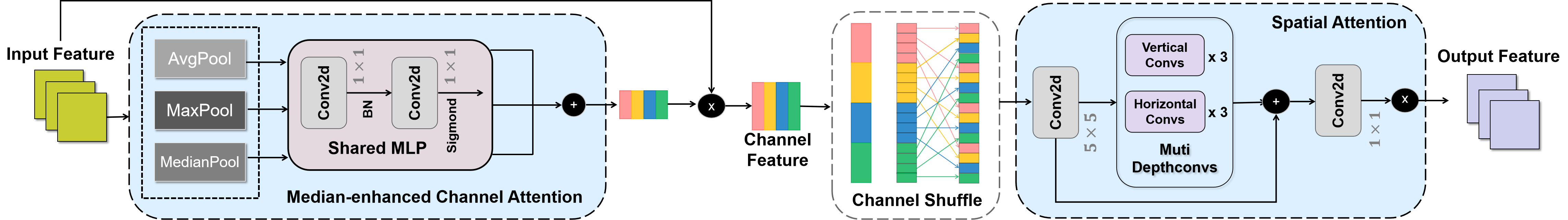}
	\caption{Illustration of the GMEA module.} 
 \label{Fig.8}
\end{figure*}

\subsection{GMEA Module}
In ISTD tasks, traditional CNNs effectively capture local features but are less robust to noise. Given that median pooling is widely used in image processing tasks to remove noise, we propose the GMEA module, which integrates a median-enhanced channel attention mechanism with a multi-scale depthwise convolution-based spatial attention, as shown in Fig. \ref{Fig.8}. A channel shuffle operation is further employed to enhance feature diversity and representation capability.

The median-enhanced channel attention block applies global average pooling (AvgPool), maximum pooling (MaxPool), and median pooling (MedPool) to the input feature map \( F \), generating three distinct descriptors. These are passed through a shared multi-layer perceptron (MLP) to produce three attention maps, which are element-wise summed to form the final channel attention map \( F_{CA} \). This map is then element-wise multiplied with \( F \) to obtain the weighted feature map \( F_{W} \). The process is defined as

\begin{equation}
	\centering
	\begin{array}{l}
    F_{CA}=\sigma(\text{MLP}(\text{AvgPool}(F))) \\[1.0ex]
    \hspace{1mm} + \sigma(\text{MLP}(\text{MaxPool}(F))) + \sigma(\text{MLP}(\text{MedPool}(F)))
	\end{array}
	\label{deqn_ex19}
\end{equation}

\begin{equation}
\label{deqn_ex20}
F_{W} = F_{CA} \odot F
\end{equation}
where \(\odot\) indicates Hadamard product.

To futher enhance information integration and feature representation, channel shuffle is applied to \(F_{W}\). The feature map is divided into four groups, each with a quarter of \( C \) channels, followed by transposition to reorder channels across groups. Finally, the shuffled feature maps \(F_{S}\) are restored to their original shape. The above process can be described as
\begin{equation}
\label{deqn_ex21}
F_{S}=\mathrm{ChannelShuffle}(F_{W})
\end{equation}

The spatial attention block uses multi-scale deep convolution, starting with an initial layer to extract basic features from the input feature map. Then, six different sizes of deep convolution layers are utilized to extract spatial feature \(F_{S}\) at different scales and directions, improving the network's perception of information across spatial scales. The outputs are element-wise summed and passed through a 1×1 convolution to generate the spatial attention map \(F_{SA}\), which is then element-wise multiplied with \(F_{SA}\) to produce the final output \(F'\). This process is formulated as
\begin{equation}
\label{deqn_ex22}
F_{SA} = \sum_{i=1}^{n} D_{i}(F_{S})
\end{equation}
\begin{equation}
\label{deqn_ex23}
F' = \mathrm{Conv_{1x1}}(F_{SA}) \odot F_{S}
\end{equation}
where \(D_{i}(\cdot)\) represents depth convolution operations of different sizes, and \(n=6\) is the number of depth convolutions.
By applying GMEA to each layer of the decoder, enhanced feature representations are achieved, enabling accurate reconstruction and restoration of small targets during decoding.

\subsection{Loss Function}
In ISTD tasks, small targets are easily obscured due to low contrast and limited size. To enhance training robustness, we adopt SoftIoU loss \cite{r54}, which directly optimizes target region overlap. It is defined as
\begin{equation}
\label{deqn_ex24}
L_{\mathrm{SoftIoU}}\left(p,y\right)=\frac{\sum_{i,j}\left(\sigma\left(p_{i,j}\right)\cdot
y_{i,j}\right)+\delta}{\sum_{i,j}\left(\sigma\left(p_{i,j}\right)+y_{i,j}-\sigma\left(p_{i,j}\right)\cdot
y_{i,j}\right)+\delta}
\end{equation}
where \(p_{i,j}\) and \(y_{i,j}\) denote the values at point \((i,j)\) on the prediction feature map and the real mask label, respectively. The smoothing factor \(\delta\) is set to 1. 

\section{EXPERIMENTAL RESULTS}
\subsection{Datasets and Implementation}
This section presents experimental validation of the proposed ARFC-WAHNet. We first introduce the datasets, evaluation metrics, and implementation details. Then, we provide comprehensive experiments using four evaluation indicators, along with quantitative and visual analyses, demonstrating that our method outperforms existing SOTA methods. Ablation studies are conducted to evaluate the contribution of each core module. Finally, key factors influencing ARFC-WAHNet’s performance and their practical implications are discussed.

\begin{table*}[t]
\caption{Comparison of detection performance [$IoU$ (\%), $F_1$ (\%), $P_d$ (\%), and $F_a$ ($\times10^{-6}$)] and model efficiency (the number of parameters (m) and theoretical flops (g)) of different methods on the SIRST, NUDT-SIRST and IRSTD-1K. The best results are in red, and the second best results are in green.}
\centering
\resizebox{\linewidth}{!}{
\begin{tabular}{l|cccccccccccccc}
\toprule
\multirow{2}{*}{Methods} & \multirow{2}{*}{Params(M)} & \multirow{2}{*}{FLOPs(G)} & \multicolumn{4}{c}{SIRST \cite{r18}} & \multicolumn{4}{c}{NUDT-SIRST \cite{r35}} & \multicolumn{4}{c}{IRSTD-1K \cite{r34}} \\
\cline{4-15}
& & & $IoU\uparrow$ & $F_1\uparrow$ & $P_d\uparrow$ & $F_a\downarrow$ & $IoU\uparrow$ & $F_1\uparrow$ & $P_d\uparrow$ & $F_a\downarrow$ & $IoU\uparrow$ & $F_1\uparrow$ & $P_d\uparrow$ & $F_a\downarrow$ \\
\hline
MPCM \cite{r8}       & -      & -      & 12.84 & 35.05 & 83.27 & 54.65 & 5.86 & 36.38 & 55.87 & 115.96 & 7.33 & 23.83 & 68.69 & 51.96 \\
IPI \cite{r6}        & -      & -      & 25.67 & 41.23 & 85.55 & 55.98 & 34.62 & 51.43 & 85.93 & 88.31 & 27.92 & 30.33 & 67.26 & 81.86 \\
PSTNN \cite{r56}     & -      & -      & 22.35 & 24.30 & 77.95 & 65.63 & 19.07 & 39.98 & 70.37 & 44.17 & 17.19 & 26.20 & 62.96 & 35.26 \\
\hline
ACM \cite{r18}       & 0.398  & 0.40   & 65.78 & 78.11 & 92.02 & 46.79 & 63.59 & 80.79 & 95.03 & 17.86 & 59.26 & 76.75 & 90.57 & 91.46 \\
ALCNet \cite{r32}    & 0.378  & 3.74   & 66.24 & 78.13 & 92.78 & 57.83 & 68.41 & 82.77 & 97.35 & 13.49 & 58.09 & 77.61 & 92.59 & 74.45 \\
DNANet \cite{r35}    & 4.697  & 14.26  & 74.03 & \textcolor{red}{87.73} & 95.06 & 62.98 & \textcolor{green!75!black}{92.39} & \textcolor{green!75!black}{93.61} & \textcolor{green!75!black}{98.52} & \textcolor{red}{3.68} & 63.72 & 77.84 & \textcolor{green!75!black}{92.93} & 48.22 \\
RDIAN \cite{r57}     & 0.0217 & 3.72   & 67.44 & 84.27 & 93.92 & 85.82 & 82.42 & 89.95 & 98.20 & 27.97 & 60.09 & 77.80 & 91.92 & 47.07 \\
UIUNet \cite{r33}    & 50.54  & 54.42  & 73.73 & 81.52 & 92.40 & \textcolor{red}{18.04} & 89.83 & 92.16 & 97.57 & \textcolor{green!75!black}{5.93} & 64.33 & 76.99 & 90.91 & \textcolor{green!75!black}{20.10} \\
AGPCNet \cite{r16}   & 12.36  & 43.18  & 73.97 & 85.04 & \textcolor{green!75!black}{96.20} & 45.36 & 85.50 & 92.18 & 97.04 & 7.28 & 64.71 & \textcolor{green!75!black}{78.72} & 89.56 & \textcolor{red}{18.56} \\
MSHNet \cite{r38}    & 4.065  & 6.11   & 72.65 & 84.16 & 90.78 & 230.9 & 76.49 & 86.67 & 96.08 & 26.40 & 64.61 & 73.73 & 87.21 & 42.26 \\
MDIGCNet \cite{r45}  & 1.505  & 6.557  & \textcolor{green!75!black}{74.21} & 72.06 & 93.65 & 56.84 & 83.88 & 85.65 & 97.04 & 13.90 & \textcolor{red}{65.20} & 77.08 & 91.25 & 32.86 \\
\rowcolor{gray!20} Ours & 3.464 & 5.86 & \textcolor{red}{74.29} & \textcolor{green!75!black}{86.41} & \textcolor{red}{96.96} & \textcolor{green!75!black}{33.08} & \textcolor{red}{93.25} & \textcolor{red}{96.05} & \textcolor{red}{98.94} & 10.04 & \textcolor{green!75!black}{64.92} & \textcolor{red}{79.21} & \textcolor{red}{93.94} & 26.61 \\
\bottomrule
\end{tabular}
}
\label{SOTA}
\vspace{-0.3cm}
\end{table*}

\begin{figure*}[t]
{
    \captionsetup[subfloat]{labelformat=parens, font=normalsize, labelfont=small} 
    \centering
    \subfloat[]{%
        \includegraphics[width=2.3in]{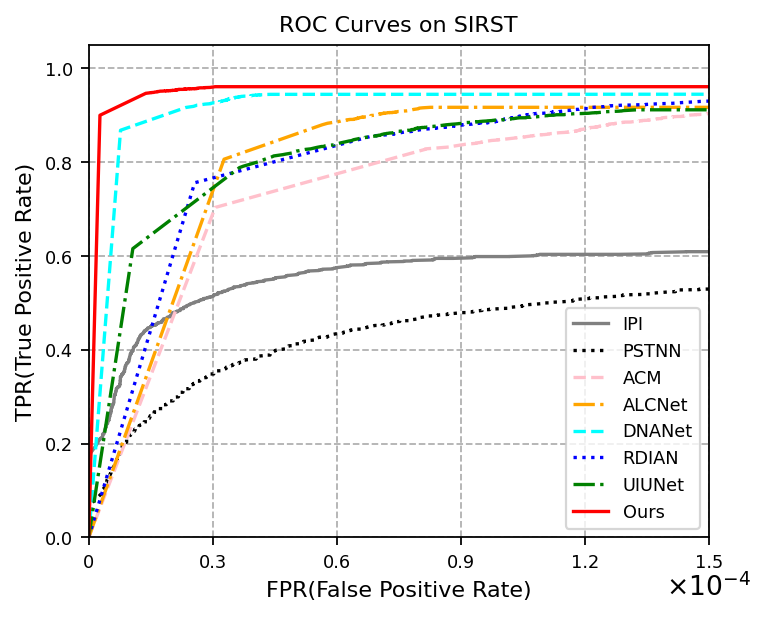}%
        \label{fig_sirst}
    }\hfil
    \subfloat[]{%
        \includegraphics[width=2.3in]{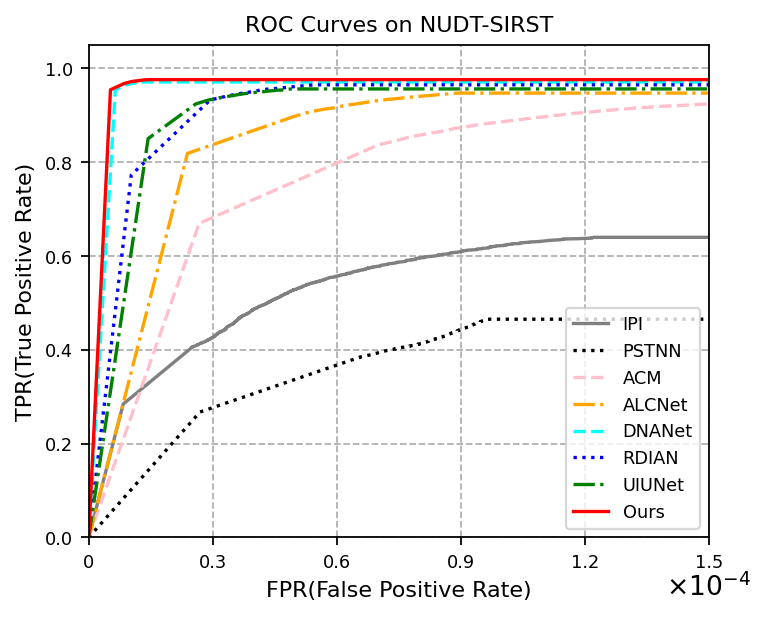}%
        \label{fig_nudt}
    }\hfil
    \subfloat[]{%
        \includegraphics[width=2.3in]{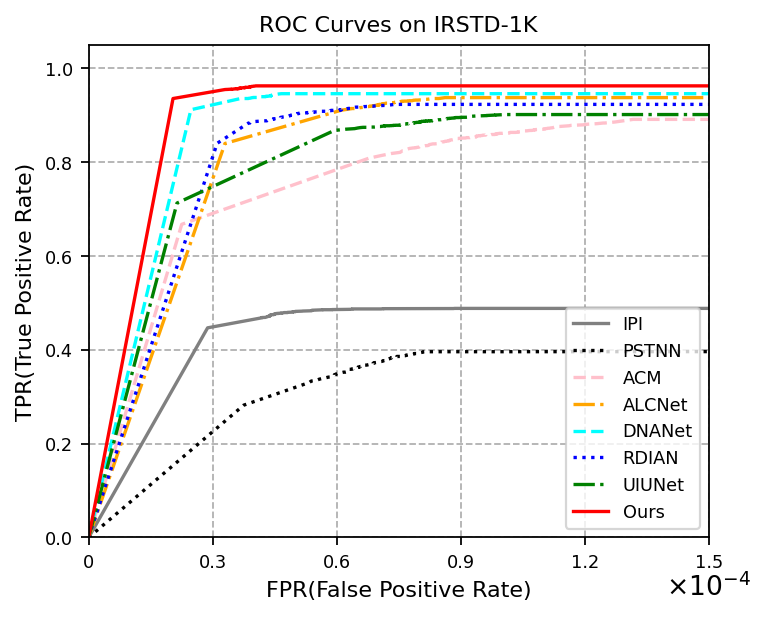}%
        \label{fig_irstd}
    }
    \caption{ROC curves of different algorithms. (a) ROC curves on SIRST. (b) ROC curves on NUDT-SIRST. (c) ROC curves on IRSTD-1K.}
    \label{ROC}
}
\end{figure*}

\subsubsection{Datasets}
We evaluate the effectiveness of the proposed ARFC-WAHNet through comparative and ablation experiments on three public benchmark datasets: SIRST \cite{r18}, NUDT-SIRST \cite{r35}, and IRSTD-1K \cite{r34}. SIRST is the first publicly available single-frame infrared small target dataset, comprising 427 images (213 for training, 214 for testing) collected from hundreds of scenarios. Approximately 90\% of the images contain a single target; 55\% of targets occupy only 0.02\% of the image area, and only 35\% are the brightest regions. NUDT-SIRST contains 1327 synthetic infrared images (256×256), with 663 for training and 664 for testing. The dataset features complex backgrounds across diverse scenes, including urban, rural, ocean, cloud, and high-brightness environments. IRSTD-1K includes 1001 real infrared images (512×512), split into 800 for training and 201 for testing. It presents significant challenges due to its diverse target sizes and shapes embedded in cluttered, complex backgrounds.

\subsubsection{Evaluation Metrics}
In this study, we employ both target-level metrics-probability of detection (\(P_d\)) and false alarm rate (\(F_a\))-and pixel-level metrics to comprehensively evaluate the performance of our method on the ISTD task. The receiver operating characteristic (ROC) curve is used to assess detection performance across varying thresholds, with the target-level threshold set to 0.5 during evaluation.

\(P_d\) measures the model’s ability to correctly detect targets and is defined as the ratio of correctly detected targets to the total number of detected targets
\begin{equation}
\label{deqn_ex25}
P_d = {\frac{T_{correct}}{T_{act}}}
\end{equation}

\(F_a\) reflects the accuracy of the detection process, calculated as the ratio of falsely predicted pixels to the total number of pixels in the image
\begin{equation}
\label{deqn_ex26}
F_a=\frac{{{P}}_{false}}{{{P}}_{all}}
\end{equation}

The \(F_1\) score offers a balanced evaluation by computing the harmonic mean of precision and recall. Precision is defined as the ratio of correctly predicted target pixels to the total predicted target pixels, while recall measures the ratio of correctly predicted target pixels to the total ground-truth target pixels. The formulation is as follows
\begin{equation}
\label{deqn_ex27}
\begin{array}{c}
\mathrm{Pre} = \frac{\mathrm{TP}}{\mathrm{TP} + \mathrm{FP}} \\[1.0ex]
\operatorname{Rec} = \frac{\mathrm{TP}}{\mathrm{TP} + \mathrm{FN}} \\[1.0ex]
F_{1} = 2 \times \frac{\mathrm{Pre} \times \mathrm{Rec}}{\mathrm{Pre} + \mathrm{Rec}}
\end{array}
\end{equation}

\(IoU\) quantifies the similarity between predicted and ground-truth targets by measuring the overlap between their segmentation regions. It is defined as

\begin{equation}
\label{deqn_ex28}
IoU =\frac{\text{intersection}}{\text{union}}
\end{equation}

The ROC curve depicts the relationship between the true positive rate (TPR) and false positive rate (FPR), where a curve closer to the point (0,1) indicates superior detection performance. TPR and FPR are defined as
\begin{equation}
\label{deqn_ex29}
\mathrm{TPR} = \frac{{\mathrm{TP}}}{{\mathrm{TP+FN}}},\mathrm{FPR} = \frac{{\mathrm{FP}}}{{\mathrm{TN+FP}}}
\end{equation}

\subsubsection{Implementation Details}
The network is trained using the Adam optimizer \cite{r55} with a MultiStepLR scheduler. Training is performed for 400 epochs with a batch size of 8 and an initial learning rate of 0.0005. The model is implemented using Python 3.10 and PyTorch 2.3.1. Baseline comparisons are conducted via the BasicISTD Toolbox\footnote{\url{https://github.com/XinyiYing/BasicIRSTD}}, while official protocols are followed for other methods. All experiments are run on a workstation with an Intel Xeon Gold 6133 CPU and an NVIDIA GeForce RTX 4090 GPU.

\begin{figure*}[t]
	\centering 
	\includegraphics[width=\textwidth]{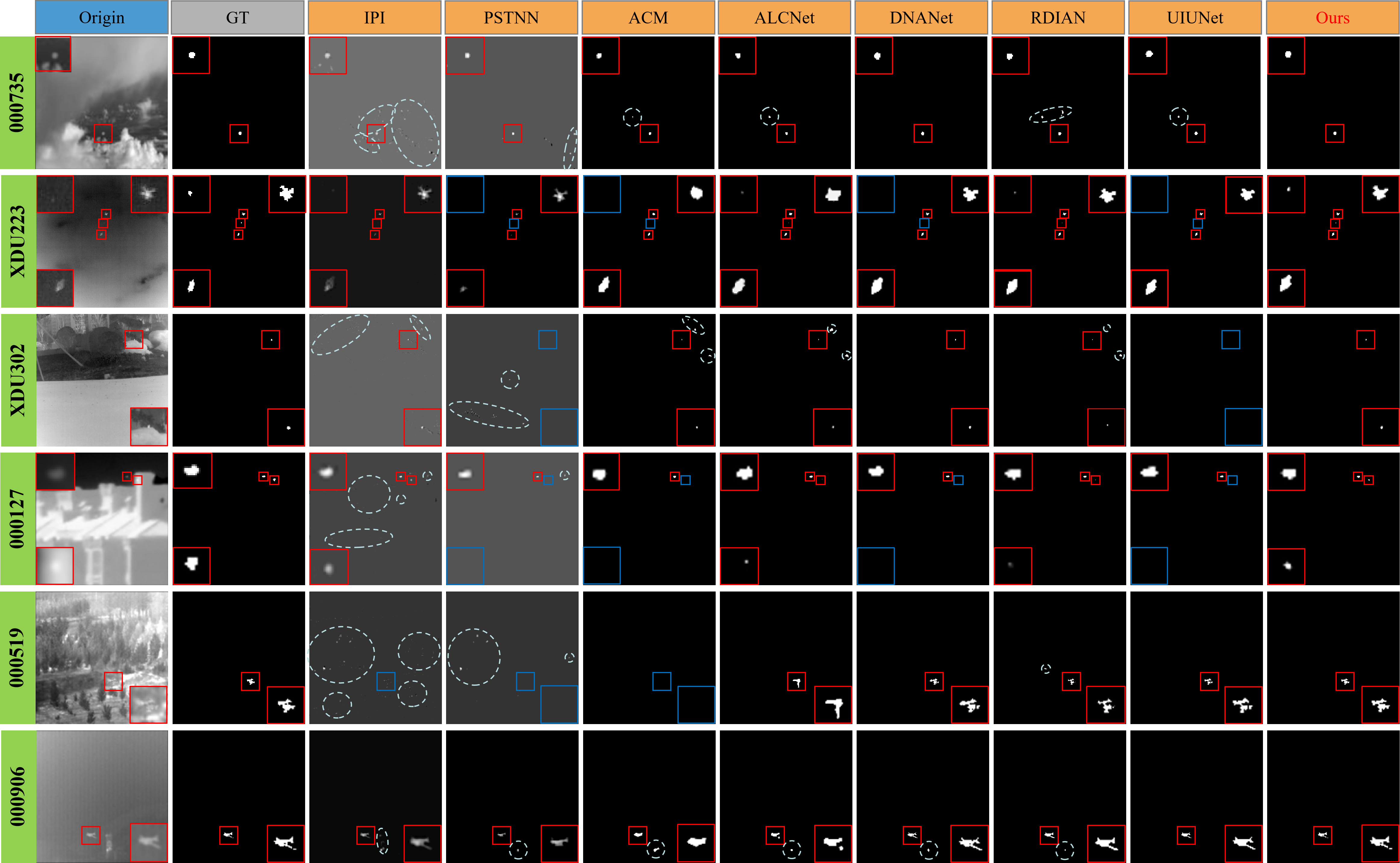}
	\caption{Visual examples of some representative methods.  In each figure, false alarms are marked with gray-green circles, and missed detections are indicated
by blue rectangles. Correctly detected targets are highlighted in red, with zoomed-in patches placed in the corners of the detection images.} 
 \label{All}
\end{figure*}

\begin{figure*}[t]
	\centering 
	\includegraphics[width=\textwidth]{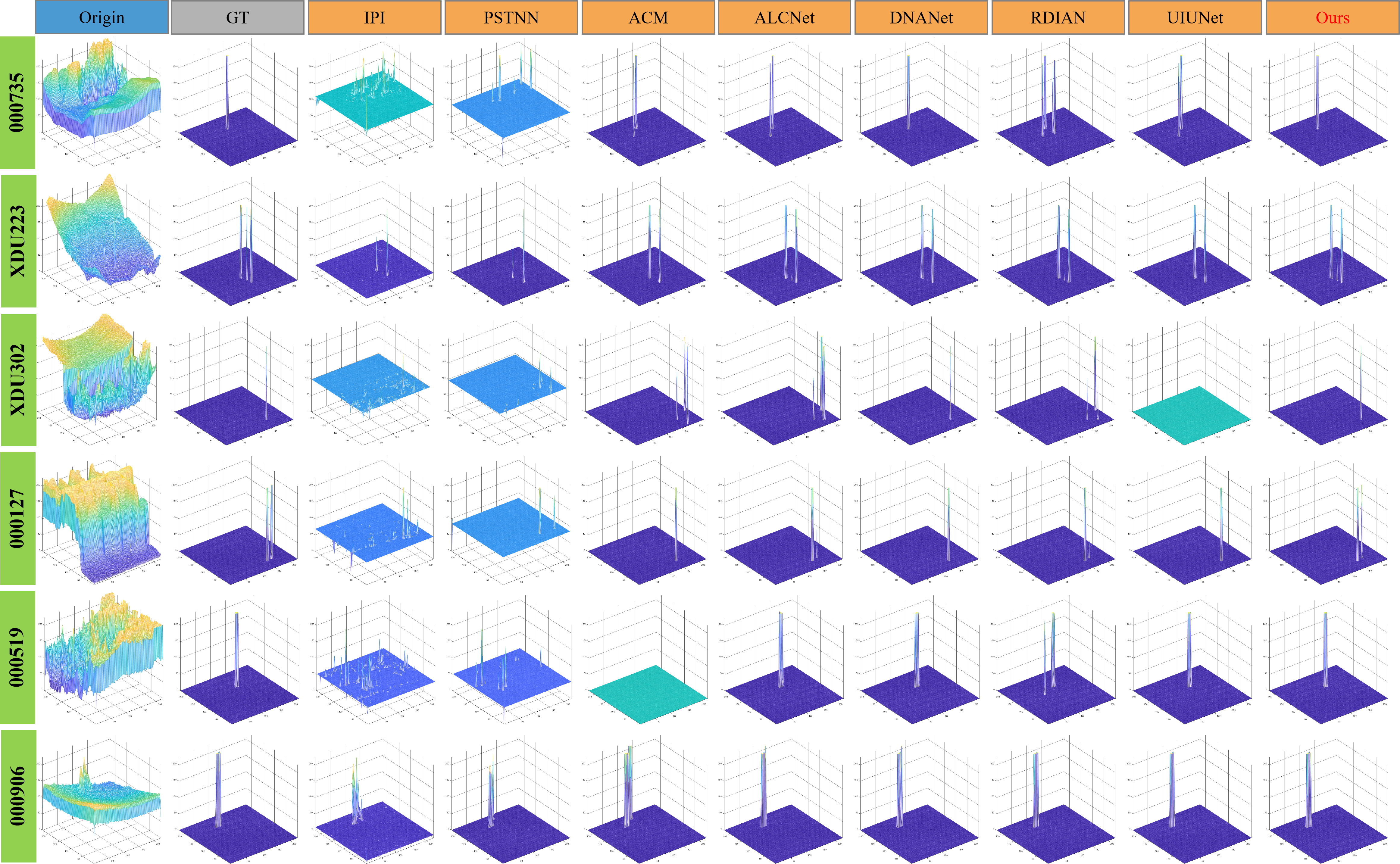}
	\caption{3D visualization results of different methods on 6 test images.} 
 \label{Fig.11}
\end{figure*}

\begin{table*}[t]
\caption{$IoU$ (\%),$F_1$(\%), $P_d$ (\%), and $F_a$ ($\times10^{-6}$) values achieved in the sirst, nudt-sirst, and irstd-1k datasets on ablation experiments about MRFFIConv. The best metrices are in bold, and the second best are underlined.}
\centering
\resizebox{\linewidth}{!}{
\begin{tabular}{l|cccccccccccccc}
\toprule
\multirow{2}{*}{Methods} & \multicolumn{4}{c}{SIRST} & & \multicolumn{4}{c}{NUDT-SIRST} & & \multicolumn{4}{c}{IRSTD-1K} \\
\cline{2-5} \cline{7-10} \cline{12-15}
& $IoU\uparrow$ & $F_1\uparrow$ & $P_d\uparrow$ & $F_a\downarrow$ 
& & $IoU\uparrow$ & $F_1\uparrow$ & $P_d\uparrow$ & $F_a\downarrow$ 
& & $IoU\uparrow$ & $F_1\uparrow$ & $P_d\uparrow$ & $F_a\downarrow$ \\
\hline
Conv2d(k=3)   & \underline{72.08} & 81.37 & 93.54 & \textbf{24.76} & & 85.50 & 88.69 & 97.46 & 8.41 & & 59.16 & \underline{75.34} & 89.56 & 43.08 \\
MSDC          & 71.10 & \underline{82.82} & \underline{94.30} & 45.41 & & \underline{88.70} & \underline{91.83} & 97.67 & \underline{6.71} & & \textbf{62.54} & \underline{76.94} & 92.59 & 57.51 \\
DCN           & 65.32 & 77.12 & 90.49 & 67.43 & & 80.03 & 83.61 & 91.25 & 13.93 & & 60.57 & 71.89 & 86.11 & 59.63 \\
MDDC          & 73.69 & 81.77 & 93.92 & \underline{37.73} & & 85.84 & \textbf{91.75} & \underline{97.88} & \textbf{5.56} & & \underline{63.28} & \textbf{76.69} & 90.91 & \underline{33.76} \\
\rowcolor{gray!20} MRFFIConv & \textbf{74.09} & \textbf{82.90} & \textbf{95.06} & 58.79 & & \textbf{89.59} & 92.12 & \textbf{98.20} & 7.72 & & 63.37 & 75.45 & \textbf{93.27} & \underline{40.63} \\
\bottomrule
\end{tabular}
}
\label{MRFFIConv_1}
\vspace{-0.3cm}
\end{table*}

\begin{table*}[t]
\caption{Comparison of quantitative metrics [$IoU$ (\%), $F_1$ (\%), $P_d$ (\%), and $F_a$ ($\times10^{-6}$)] for the application of MRFFIConv on other networks (MSHNet, DNANet, and UIUNet).}
\centering
\resizebox{\linewidth}{!}{
\begin{tabular}{l|c|cccccccccccccc}
\toprule
\multirow{2}{*}{Methods} & \multirow{2}{*}{MRFFIConv} & \multicolumn{4}{c}{SIRST} & & \multicolumn{4}{c}{NUDT-SIRST} & & \multicolumn{4}{c}{IRSTD-1K} \\
\cline{3-6} \cline{8-11} \cline{13-16}
& & $IoU\uparrow$ & $F_1\uparrow$ & $P_d\uparrow$ & $F_a\downarrow$
& & $IoU\uparrow$ & $F_1\uparrow$ & $P_d\uparrow$ & $F_a\downarrow$
& & $IoU\uparrow$ & $F_1\uparrow$ & $P_d\uparrow$ & $F_a\downarrow$ \\
\hline
\multirow{2}{*}{MSHNet}
& \ding{53} & 72.65 & \textbf{84.16} & 90.78 & 230.9 & & 76.49 & 86.67 & 96.08 & 26.40 & & 64.61 & \textbf{73.73} & \textbf{87.21} & 42.26 \\
& \ding{52} & \textbf{75.73} & 83.81 & \textbf{93.16} & \textbf{170.49} & & \textbf{83.21} & \textbf{88.45} & \textbf{96.93} & \textbf{24.70} & & \textbf{66.05} & 72.06 & 86.67 & \textbf{23.19} \\
\multirow{2}{*}{DNANet}
& \ding{53} & 74.03 & 87.73 & 95.06 & 62.98 & & 92.39 & 93.61 & 98.52 & \textbf{3.68} & & 63.72 & 77.84 & 92.93 & 48.22 \\
& \ding{52} & \textbf{77.16} & \textbf{88.41} & \textbf{95.44} & \textbf{54.26} & & \textbf{93.97} & \textbf{96.31} & \textbf{98.94} & 4.23 & & \textbf{64.52} & \textbf{78.36} & \textbf{93.27} & \textbf{36.50} \\
\multirow{2}{*}{UIUNet}
& \ding{53} & 73.73 & 81.52 & 92.40 & \textbf{18.04} & & 89.83 & 92.16 & 97.57 & 5.93 & & \textbf{64.33} & 76.99 & \textbf{90.91} & \textbf{20.10} \\
& \ding{52} & \textbf{74.49} & \textbf{83.19} & \textbf{93.92} & 31.49 & & \textbf{92.41} & \textbf{96.03} & \textbf{97.78} & \textbf{4.18} & & 63.28 & \textbf{77.37} & 90.57 & 25.49 \\
\bottomrule
\end{tabular}
}
\label{MRFFIConv_2}
\vspace{-0.3cm}
\end{table*}

\begin{table*}[t]
\caption{$IoU$ (\%), $F_1$ (\%), $P_d$ (\%), and $F_a$ ($\times10^{-6}$) values achieved in the SIRST, NUDT-SIRST, and IRSTD-1K datasets on ablation experiments about WFED. The best metrics are in bold, and the second best are underlined.}
\centering
\resizebox{\linewidth}{!}{
\begin{tabular}{l|cccccccccccccc}
\toprule
\multirow{2}{*}{Methods} & \multicolumn{4}{c}{SIRST} & & \multicolumn{4}{c}{NUDT-SIRST} & & \multicolumn{4}{c}{IRSTD-1K} \\
\cline{2-5} \cline{7-10} \cline{12-15}
& $IoU\uparrow$ & $F_1\uparrow$ & $P_d\uparrow$ & $F_a\downarrow$
& & $IoU\uparrow$ & $F_1\uparrow$ & $P_d\uparrow$ & $F_a\downarrow$
& & $IoU\uparrow$ & $F_1\uparrow$ & $P_d\uparrow$ & $F_a\downarrow$ \\
\hline
MaxPool       & 72.08  & 81.37  & 93.54  & \textbf{24.76} & & 85.50  & 88.69  & 97.46  & 8.41   & & 59.16  & 75.34  & 89.56  & 43.08  \\
AvgPool       & 65.78  & 79.22  & 92.40  & 38.69  & & 81.43  & 80.20  & 95.03  & 19.14  & & \underline{62.07}  & 72.91  & 89.26  & 48.22  \\
DWT           & 69.91  & 81.47  & \textbf{94.67} & 50.90  & & 86.90  & 90.98  & \underline{97.67}  & 12.77  & & 61.67  & 75.16  & 90.24  & 32.70  \\
HWT           & \underline{73.46}  & \underline{82.13}  & \underline{94.30}  & 31.49  & & \underline{87.87}  & \textbf{91.82}  & 97.35  & \textbf{5.56}  & & 61.24  & \underline{75.01}  & \underline{91.25}  & \textbf{29.13} \\
\rowcolor{gray!20} WFED & \textbf{74.19} & \textbf{83.81} & \textbf{94.67} & \underline{27.44} & & \textbf{88.80} & \underline{91.76}  & \textbf{97.78} & \underline{5.63}  & & \textbf{64.69} & \textbf{75.92} & \textbf{91.92} & \underline{31.73} \\
\bottomrule
\end{tabular}
}
\label{WFED}
\vspace{-0.3cm}
\end{table*}

\subsection{Comparison to SOTA Methods}
In this section, we compared our methods with SOTA ISTD methods in model-driven (MPCM \cite{r8}, IPI \cite{r6}, and PSTNN \cite{r56}), and data-driven (ACM \cite{r18}, ALCNet \cite{r32}, DNANet \cite{r35}, RDIAN \cite{r57}, UIUNet \cite{r33}, AGPCNet \cite{r16} ,MSHNet \cite{r38}, and MDIGCNet \cite{r45}) on three public datasets.

\subsubsection{Qualitative Results}
Table \ref{SOTA} presents the quantitative results on SIRST, NUDT-SIRST, and IRSTD-1K. The best performance for each metric is highlighted in red, and the second-best in green. Overall, data-driven methods generally outperform model-driven ones, though performance varies across datasets. MPCM as a classic ISTD approach, achieves relatively high \(P_d\) but performs the worst in \(IoU\) and \(F_1\), indicating limited detail preservation. Model-driven methods such as IPI and PSTNN demonstrate moderate improvements in \(IoU\) and \(F_1\); however, their reliance on hand-crafted parameters limits their generalization to complex scenarios, and their detection performance may degrade when multiple targets are present in a single image. Among data-driven approaches, ACM performs the worst across all datasets, while MSHNet struggles in complex backgrounds. ALCNet, RDIAN, and MDIGCNet offer improvements over traditional methods but remain average among deep learning models. In contrast, DNANet ranks second in segmentation performance on both NUDT-SIRST and IRSTD-1K, while AGPCNet achieves the second-best \(P_d\) and \(F_1\) scores on SIRST. UIUNet delivers consistently good performance across datasets but suffers from the highest Params and FLOPs among all evaluated methods.

The proposed ARFC-WAHNet achieves strong performance with relatively low- computational overhead, attaining the best results in both \(P_d\) and \(F_1\). Although its \(F_a\) performance isn't the highest across all datasets, it ranks second on SIRST and third on IRSTD-1K, indicating competitive background suppression.  Compared to lightweight methods like ALCNet and RDIAN, our ARFC-WAHNet achieves superior performance across all metrics on the three datasets. On SIRST, our method improves \(P_d\) by approximately 4\%. Among similarly efficient models such as DNANet, MSHNet, and MDIGCNet, our approach consistently achieves either the best or second-best results. Moreover, even against models with significantly higher parameter counts and FLOPs-such as UIUNet and AGPCNet-our ARFC-WAHNet outperforms them in pixel-level metrics, demonstrating more effective feature utilization, highlighting its favorable trade-off between performance and efficiency, and thus offering higher practical applicability.

Fig. \ref{ROC} shows the ROC curves comparing our method with other SOTA approaches on SIRST, NUDT-SIRST, and IRSTD-1K. ARFC-WAHNet achieves the highest \(P_d\) and \(F_1\) on both the SIRST and NUDT-SIRST, demonstrating robust detection capability. In particular, our network maintains an excellent balance between \(P_d\) and \(F_a\), ensuring stable performance across scenes. In contrast, model-driven approaches like IPI and PSTNN show inconsistent results across datasets, reflecting limited robustness in complex infrared scenarios.

\subsubsection{Visual Comparison}
As shown in Fig. \ref{All}, we visualize six representative scenes selected from the three datasets, including sky, urban, and forest environments, with either single or multiple targets. In each image, false alarms are marked with gray-green circles, and missed detections are indicated by blue rectangles. Correctly detected targets are highlighted in red, with zoomed-in patches placed in the corners of the detection images. Fig. \ref{Fig.11} shows the 3D visualization results of different methods on 6 test images. Traditional model-driven methods like IPI and PSTNN, often suffer from numerous false alarms or missed detections, and tend to localize only the approximate positions of targets. This issue becomes more prominent in cases with low target-to-background contrast, such as NUDT-SIRST 000519 and IRSTD-1K XDU302. Such performance degradation mainly stems from their reliance on assumptions of sparsity and low-rank decomposition, which limits their effectiveness in complex scenarios compared to deep learning-based approaches.

Compared with other deep learning-based methods, the proposed ARFC-WAHNet achieves the lowest miss rate while accurately detecting targets and preserving their fine structural details. It outperforms ACM, ALCNet, and RDIAN with fewer false alarms and more precise boundary delineation, especially in scenes with complex patterns (e.g., NUDT-SIRST 000127 and 000906). It also surpasses DNANet and UIUNet in multi-target scenarios such as IRSTD-1K XDU223. These advantages stem from MRFFIConv, which enhances scene adaptability and edge feature extraction, improving small target shape recovery. The WFED module suppresses background noise while highlighting target structures, reducing omission risks. Finally, HLFF and GMEA jointly enhance feature fusion and utilization, further minimizing missed detections.

\begin{table*}[t]
\caption{$IoU$ (\%), $F_1$ (\%), $P_d$ (\%), and $F_a$ ($\times10^{-6}$) values achieved in the SIRST, NUDT-SIRST, and IRSTD-1K datasets on ablation experiments about HLFF. The best metrics are in bold, and the second best are underlined.}
\centering
\resizebox{\linewidth}{!}{
\begin{tabular}{l|cccccccccccccc}
\toprule
\multirow{2}{*}{Methods} & \multicolumn{4}{c}{SIRST} & & \multicolumn{4}{c}{NUDT-SIRST} & & \multicolumn{4}{c}{IRSTD-1K} \\
\cline{2-5} \cline{7-10} \cline{12-15}
& $IoU\uparrow$ & $F_1\uparrow$ & $P_d\uparrow$ & $F_a\downarrow$
& & $IoU\uparrow$ & $F_1\uparrow$ & $P_d\uparrow$ & $F_a\downarrow$
& & $IoU\uparrow$ & $F_1\uparrow$ & $P_d\uparrow$ & $F_a\downarrow$ \\
\hline
Backbone                & 72.08  & 81.37  & 93.54  & \textbf{24.76}  & & 85.50  & 88.69  & 97.46  & 8.41   & & 59.16  & 75.34  & 89.56  & 43.08  \\
HLFF                    & 72.05  & 81.59  & 94.30  & 28.61  & & 88.49  & 92.59  & 97.67  & \underline{8.00}   & & 61.16  & \underline{76.95}  & 91.58  & 43.04  \\
MRFFIConv+HLFF          & 72.75  & 82.20  & \underline{95.06}  & 34.85  & & \underline{89.57}  & \underline{95.00}  & \underline{98.10}  & 8.07   & & \textbf{64.35} & \textbf{77.70} & 92.59  & \textbf{33.76} \\
WFED+HLFF               & \underline{73.71} & \textbf{87.46} & 93.92  & 37.87  & & 88.51  & 93.13  & 97.99  & 15.10  & & \underline{63.31}  & 76.13  & \underline{92.93}  & 40.90  \\
\rowcolor{gray!20} MRFFIConv+WFED+HLFF & \textbf{73.93} & \underline{83.31} & \textbf{95.44} & \underline{28.47} & & \textbf{91.66} & \textbf{95.63} & \textbf{98.42} & \textbf{4.57} & & 62.43  & 76.88  & \textbf{93.60} & \textbf{30.01} \\
\bottomrule
\end{tabular}
}
\label{HLFF}
\vspace{-0.3cm}
\end{table*}

\begin{figure}[t]
	\centering 
	\includegraphics[width=0.49\textwidth]{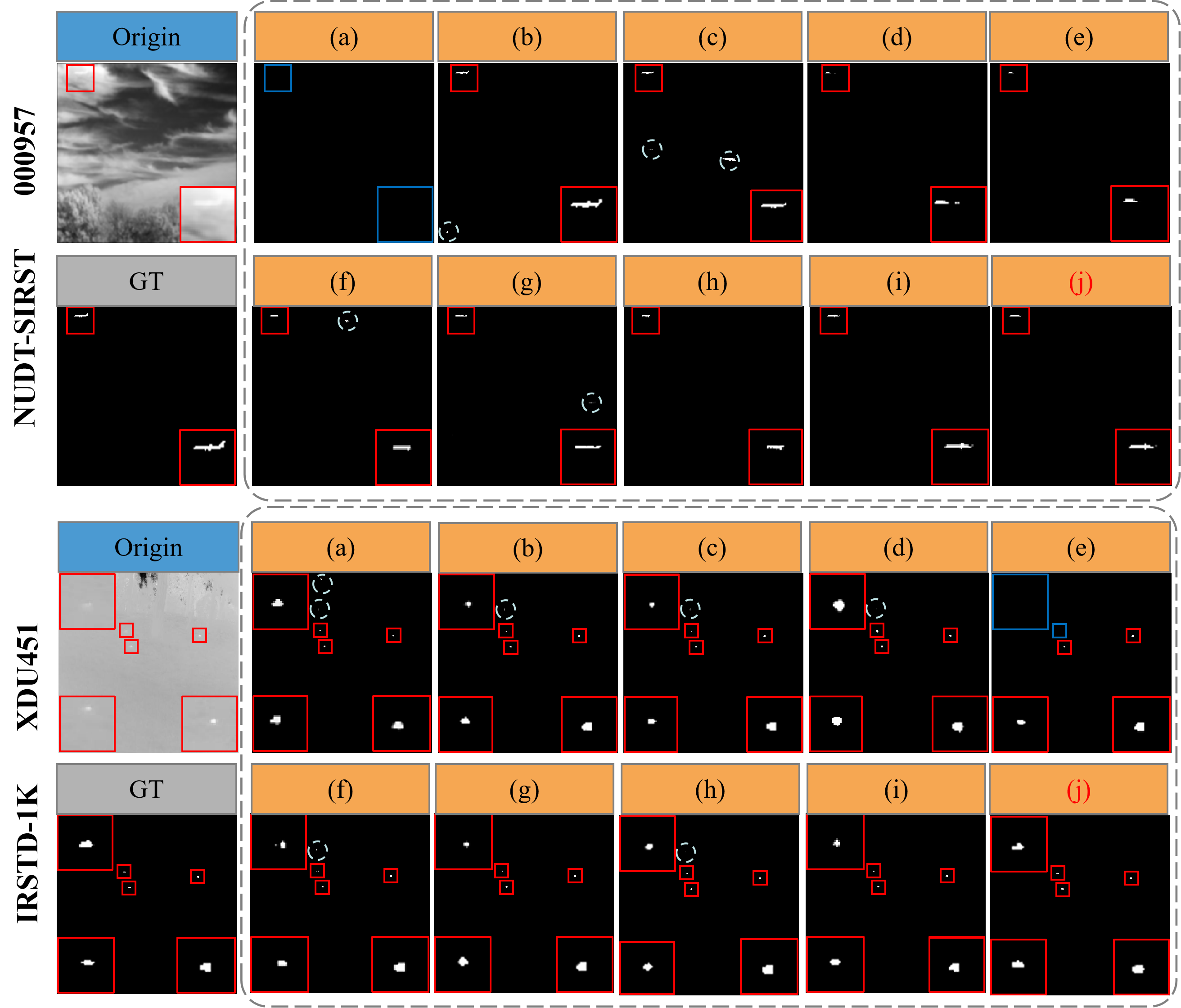}
	\caption{Results of ablation experiment. (a) Detection results of backbone. (b) Detection results of backbone \& MRFFIConv. (c) Detection results of backbone \& WFED. (d) Detection results of backbone \& HLFF. (e) Detection results of backbone \& GMEA. (f) Detection results without MRFFIConv. (g) Detection results without WFED. (h) Detection results without HLFF. (i) Detection results without GMEA. (j) Detection results of ARFC-WAHNet.} 
 \label{Fig.12}
\end{figure}\textbf{}

\begin{table*}[t]
\caption{$IoU$ (\%), $F_1$ (\%), $P_d$ (\%), and $F_a$ ($\times10^{-6}$) values achieved in the SIRST, NUDT-SIRST, and IRSTD-1K datasets on ablation experiments about GMEA. The best metrics are in bold, and the second best are underlined.}
\centering
\resizebox{\linewidth}{!}{
\begin{tabular}{l|cccccccccccccc}
\toprule
\multirow{2}{*}{Methods} & \multicolumn{4}{c}{SIRST} & & \multicolumn{4}{c}{NUDT-SIRST} & & \multicolumn{4}{c}{IRSTD-1K} \\
\cline{2-5} \cline{7-10} \cline{12-15}
& $IoU\uparrow$ & $F_1\uparrow$ & $P_d\uparrow$ & $F_a\downarrow$
& & $IoU\uparrow$ & $F_1\uparrow$ & $P_d\uparrow$ & $F_a\downarrow$
& & $IoU\uparrow$ & $F_1\uparrow$ & $P_d\uparrow$ & $F_a\downarrow$ \\
\hline
Backbone                & 72.08  & 81.37  & 93.54  & 24.76  & & 85.50  & 88.69  & 97.46  & 8.41   & & 59.16  & 75.34  & 89.56  & 43.08  \\
GMEA                    & 73.06  & 83.56  & 94.30  & 29.22  & & 88.70  & 88.78  & 97.78  & 16.66  & & 59.45  & 75.46  & 90.57  & \textbf{17.19} \\
MRFFIConv+GMEA          & 73.04  & 84.42  & \underline{95.82}  & 41.85  & & \textbf{93.25} & 95.25  & \underline{98.52}  & \underline{3.48}   & & 62.74  & 74.58  & \underline{91.92}  & 50.50  \\
WFED+GMEA               & \underline{73.72}  & \textbf{86.20} & 94.68  & \textbf{23.67} & & \underline{93.11}  & 91.74  & 98.41  & 4.02   & & 62.10  & 76.64  & 91.58  & \underline{30.40}  \\
HLFF+ GMEA              & 73.64  & 84.19  & 95.44  & 37.52  & & 92.99  & \textbf{96.48} & 98.20  & \textbf{3.40}   & & \underline{62.37}  & \underline{77.01}  & 91.25  & 30.52  \\
\rowcolor{gray!20} MRFFIConv+WFED+GMEA & \textbf{73.95} & \underline{85.84}  & \textbf{96.20} & \underline{24.30}  & & 91.31  & \underline{96.43}  & \textbf{99.05} & 12.20  & & \textbf{62.83} & \textbf{77.19} & \textbf{92.26} & 31.26  \\
\bottomrule
\end{tabular}
}
\label{GMEA}
\vspace{-0.3cm}
\end{table*}

\begin{figure}[t]
	\centering 
	\includegraphics[width=0.49\textwidth]{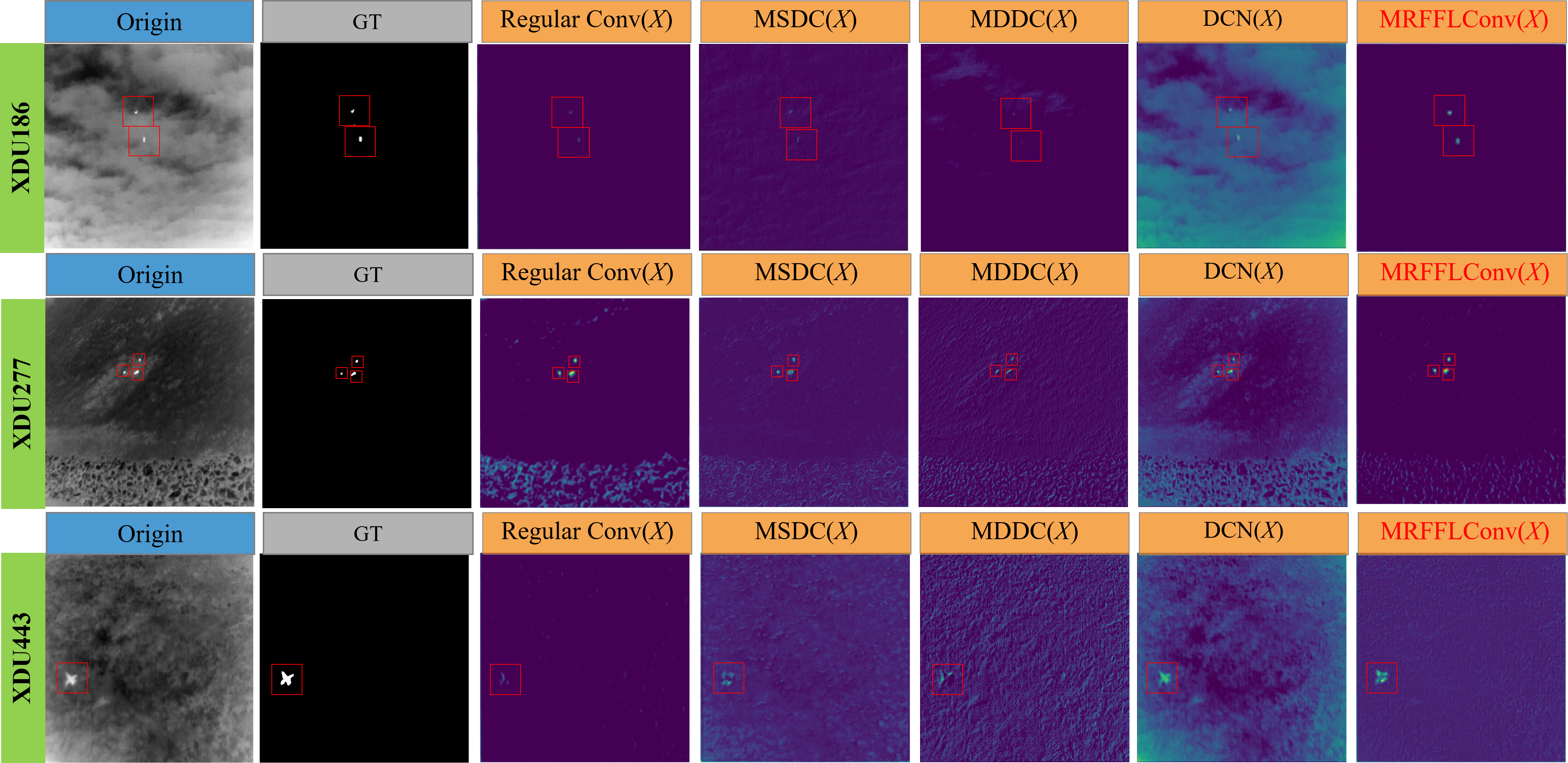}
	\caption{Illustration of heatmap. The columns from left to right represent the original image, ground truth, heatmap output from Regular Conv, MSDC, MDDC, DCN, and MRFFLConv, respectively.} 
 \label{Fig.13}
\end{figure}\textbf{}

\subsection{Ablation Study}
To evaluate each module’s contribution, we conduct ablation studies on SIRST, NUDT-SIRST, and IRSTD-1K under identical settings. Results are summarized in the tables, with the best values highlighted in bold. As shown in Fig. \ref{Fig.12}, visualizations further confirm the effectiveness of individual modules and their combination.

\subsubsection{Effect of MRFFIConv}
We compare the performance of the backbone network with and without the MRFFIConv module and further evaluate the effect of its individual convolution branches: MSDC, DCN, and MDDC. As shown in Table \ref{MRFFIConv_1}, the model with MRFFIConv achieves the best results on SIRST and NUDT-SIRST, especially in terms of \(IoU\), \(F_1\), and \(P_d\) . It also performs competitively on IRSTD-1K. Among the three branches, MSDC and MDDC perform better in most cases, as they are more effective for infrared small targets, which often have low contrast and require multi-scale feature extraction. In contrast, DCN alone shows weaker generalization, likely due to the limited presence of irregular-shaped targets in the datasets. However, it still contributes to shape-adaptive enhancement and is therefore retained in the final design. The heatmaps in Fig. \ref{Fig.13} further illustrate the complementary strengths of the three branches. While each performs well in specific scenarios, MRFFIConv consistently enhances target features and suppresses background interference, making it effective across different infrared scenes.

In addition, to further verify the robustness and effectiveness of MRFFIConv, we integrate it into three existing networks: MSHNet, DNANet, and UIUNet, by replacing their standard 3×3 convolutions. As shown in Table \ref{MRFFIConv_2}, most metrics on the SIRST and NUDT-SIRST datasets show improvements after this substitution. Although performance on IRSTD-1K is slightly lower, DNANet achieves improvements on all metrics, while MSHNet and UIUNet show significant gains in \(F_a\) and \(F_1\), respectively. These results confirm that MRFFIConv is a robust and effective module that generalizes well across different ISTD networks.

\subsubsection{Effect of WFED}
We compare the proposed WFED module with other downsampling methods, including MaxPool, AvgPool, Daubechies Wavelet Transform (DWT), and Haar Wavelet Transform (HWT). As shown in Table \ref{WFED}, MaxPool performs better than AvgPool for this task, but both are less effective than wavelet-based methods. To evaluate the impact of wavelet type, we train and test the network using Haar and Daubechies wavelets. While both yield similar performance, Haar is chosen for its simpler structure and higher computational efficiency. Among all methods, the proposed WFED achieves the best results on all three datasets. Specifically, \(IoU\) improves by 2.09\%, 3.30\%, and 5.53\%, while \(P_d\) increases by 1.13\%, 0.32\%, and 2.36\% on SIRST, NUDT-SIRST, and IRSTD-1K, respectively. Overall, these results confirm that incorporating WFED significantly enhances model performance across different datasets.

\subsubsection{Effect of HLFF}
As shown in Table \ref{HLFF}, the addition of the HLFF module consistently improves performance over the baseline across all three datasets, confirming its effectiveness. When combined with MRFFIConv, WFED, or both, HLFF achieves the best overall results on NUDT-SIRST. Although \(IoU\) and \(F_1\) slightly decrease on SIRST and IRSTD-1K compared to using MRFFIConv and WFED alone, \(P_d\) and \(F_a\) improve, indicating that HLFF complements other modules and enhances overall performance.

\subsubsection{Effect of GMEA}
As shown in Table \ref{GMEA}, the results demonstrate the impact of the GMEA module on performance. Compared with the baseline network, GMEA yields notable improvements on the IRSTD-1K dataset, with \(IoU\), \(F_1\), and \(P_d\) increasing by 0.29\%, 0.12\%, and 1.01\%, respectively, while \(F_a\) is reduced to less than half of the baseline. On SIRST and NUDT-SIRST, although \(F_a\) shows slight degradation, the other three metrics improve, confirming the effectiveness of the GMEA module. Compared with networks using one or more of the other three proposed modules, adding GMEA consistently achieves the best or second-best results across all datasets, indicating its strong complementarity and overall contribution to performance.


\section{CONCLUSION}
We propose ARFC-WAHNet, a novel framework for infrared small target detection, designed to address challenges posed by complex backgrounds, diverse target characteristics, and information loss in existing methods. To this end, we produce four specialized modules: MRFFIConv, embedded in the encoder-decoder backbone, combines multi-branch convolutions with a dynamic gated unit to enable adaptive feature extraction across varying target distributions; WFED enhances fine-grained structures and suppresses background clutter during downsampling, improving edge and detail preservation; HLFF, applied in skip connections, fuses low-level detail with high-level semantics to bridge feature hierarchies; and GMEA introduces global statistical attention to enrich feature diversity and strengthen representation capacity. Extensive experiments on benchmark datasets demonstrate that ARFC-WAHNet outperforms recent SOTA methods, achieving higher precision and lower false alarm rates, especially in challenging ISTD scenarios, while maintaining strong visual quality.

Although ARFC-WAHNet performs well on standard datasets, it still faces challenges in real-world applications, including real-time demands, background clutter, and low-contrast targets. Future work will explore structural re-parameterization to reduce model complexity and extend ARFC-WAHNet to datasets with ultra-small targets (even single-pixel level), aiming to improve robustness and reduce false alarms in complex scenes.

\bibliographystyle{IEEEtran}
\bibliography{ARFC-WAHNet}

\begin{IEEEbiography}[{\includegraphics[width=1in,height=1.25in,clip,keepaspectratio]{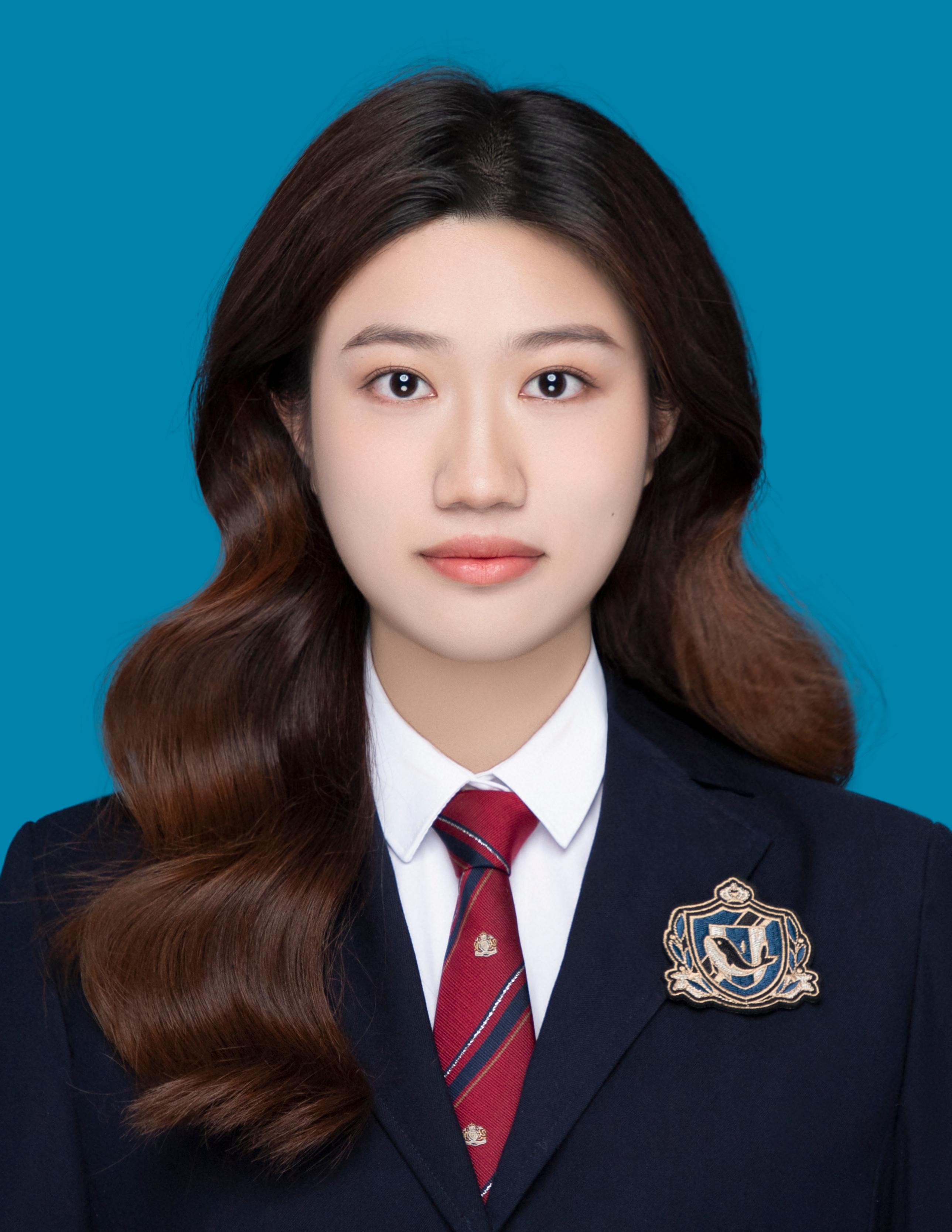}}]{Xingye Cui} received the B.E. degree in communication engineering from the School of Computer Science, Jiangsu University of Science and Technology, Zhenjiang, China, in 2023. She is currently working toward the M.E. degree in electronic information with the School of Information and Communication Engineering, University of Electronic Science and Technology of China, Chengdu, China. Her research interests include image processing, computer vision, and infrared small target detection.
\end{IEEEbiography}

\begin{IEEEbiography}[{\includegraphics[width=1in,height=1.25in,clip,keepaspectratio]{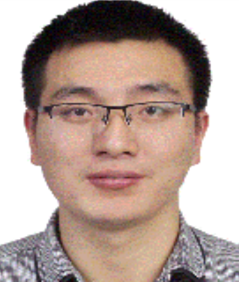}}]{Junhai Luo}
(Member, IEEE, and CCF) received a B.S. degree in computer science and appliance from the University of Electronic Science and Technology of China in 2003, an M.S. degree in computer appliance technology from the Chengdu University of Technology, Chengdu, China, in 2006, and a Ph.D. degree in information and communication engineering from the University of Electronic Science and Technology of China. He was a visiting scholar at McGill University, Canada, and the University of Tennessee, Knoxville, TN, USA. He was promoted to Associate Professor in 2011. His research interests and papers are primarily in target detection and information fusion.
\end{IEEEbiography}

\begin{IEEEbiography}[{\includegraphics[width=1in,height=1.25in,clip,keepaspectratio]{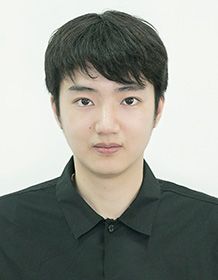}}]{Jiakun Deng} received the M.S. degree in optical engineering from the School of Optoelectronic Science and Engineering, University of Electronic Science and Technology of China, Chengdu, China, in 2022. He is currently pursuing the Ph.D. degree in Electronic Information with the School of Information and Communication Engineering, University of Electronic Science and Technology of China. His research interests include computer vision, infrared small target detection and tracking, and object recognition.
\end{IEEEbiography}

\begin{IEEEbiography}[{\includegraphics[width=1in,height=1.25in,clip,keepaspectratio]{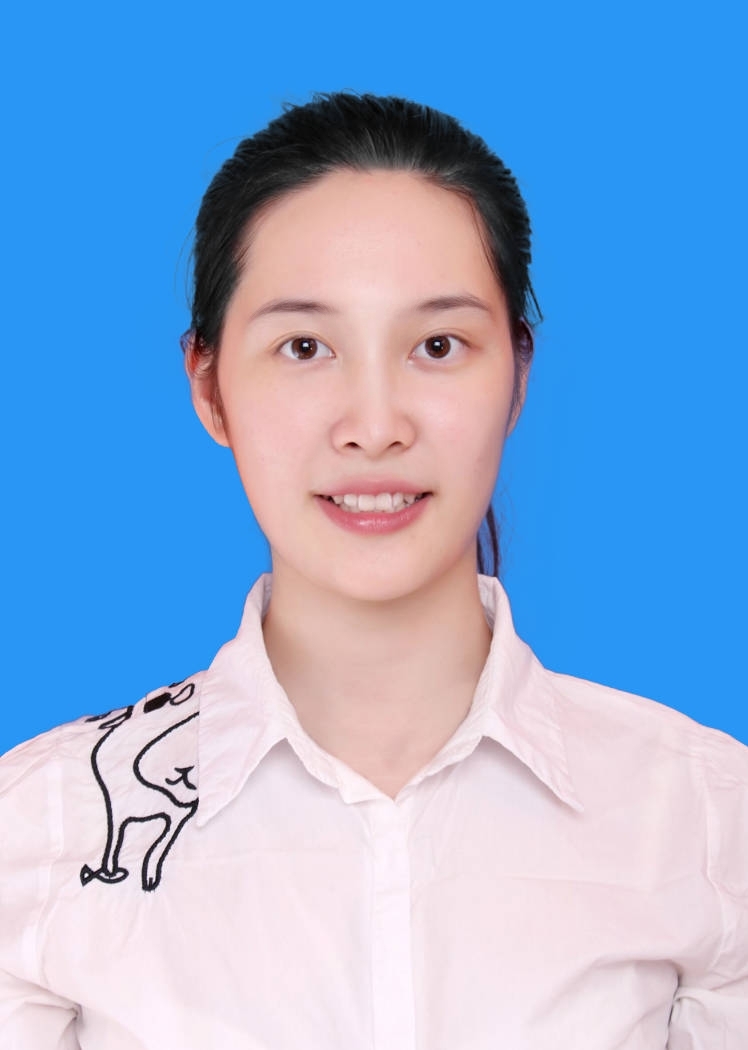}}]{Kexuan Li} received the B.S. degree from Huaqiao University, Quanzhou, China in 2023. She is currently pursuing the M.E. degree with University of Electronic Science and Technology of China, Chengdu, China. Her research interests include image processing, computer vision, and target recognition.
\end{IEEEbiography}

\begin{IEEEbiography}[{\includegraphics[width=1in,height=1.25in,clip,keepaspectratio]{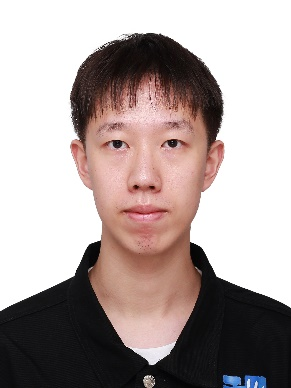}}]{Xiangyu Qiu} (Student Member, IEEE) received his B.E. degree from the school of Information and Communication Engineering, University of Electronic Science and Technology of China, in 2024. He is pursuing an M.E. degree in School of Information and Communication Engineering, University of Electronic Science and Technology of China. His research interests include image processing, computer vision, and infrared small target detection.
\end{IEEEbiography}

\vspace{-0.3cm}
\begin{IEEEbiography}[{\includegraphics[width=1in,height=1.25in,clip,keepaspectratio]{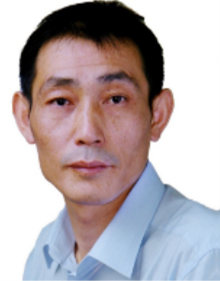}}]{Zhenming Peng}
(Member, IEEE) received his Ph.D. degree in geodetection and information technology from the Chengdu University of Technology, Chengdu, China, in 2001. From 2001 to 2003, he was a post-doctoral researcher with the Institute of Optics and Electronics, Chinese Academy of Sciences, Chengdu, China. He is currently a Professor with the University of Electronic Science and Technology of China, Chengdu. His research interests include image processing, machine learning, objects detection and remote sensing applications. Prof. Peng is members of many academic organizations, such as Institute of Electrical and Electronics Engineers (IEEE), Optical Society of America (OSA), China Optical Engineering Society (COES), Chinese Association of Automation (CAA), Chinese Society of Astronautics (CSA), Chinese Institute of Electronics (CIE), and China Society of Image and Graphics (CSIG), etc.
\end{IEEEbiography}

 




\vspace{11pt}
\end{document}